\documentclass[10pt,twocolumn,letterpaper]{article}
\usepackage[pagenumbers]{cvpr}
\usepackage{amsmath,amssymb,amsfonts}
\usepackage{textcomp}
\usepackage{xcolor}
\usepackage{multirow}
\usepackage{tabularx}
\usepackage{threeparttable}
\usepackage{hhline}
\usepackage{array}
\usepackage{booktabs}
\usepackage{xfrac}
\usepackage{algorithm}
\usepackage{algorithmicx}
\usepackage{algpseudocode}
\usepackage{bbding}

\def\BibTeX{{\rm B\kern-.05em{\sc i\kern-.025em b}\kern-.08em
    T\kern-.1667em\lower.7ex\hbox{E}\kern-.125emX}}

\newcolumntype{P}[1]{>{\centering\arraybackslash}p{#1}}
\newcolumntype{L}[1]{>{\raggedright\let\newline\\\arraybackslash\hspace{0pt}}m{#1}}
\newcolumntype{C}[1]{>{\centering\let\newline\\\arraybackslash\hspace{0pt}}m{#1}}
\newcolumntype{R}[1]{>{\raggedleft\let\newline\\\arraybackslash\hspace{0pt}}m{#1}}

\newsavebox{\compacttablebox}
\newcommand{\fitcompacttable}[2][\columnwidth]{%
  \sbox{\compacttablebox}{#2}%
  \ifdim\wd\compacttablebox>#1
    \resizebox{#1}{!}{\usebox{\compacttablebox}}%
  \else
    \usebox{\compacttablebox}%
  \fi
}
\newcommand{\fitwidetable}[1]{\fitcompacttable[\textwidth]{#1}}

\newcommand{\methodname}{SAS~Tokens}

\definecolor{vscodegreen}{RGB}{78,154,6}

\providecommand{\DIFdel}[1]{}

\providecommand{\DIFdelFL}[1]{}

\definecolor{cvprblue}{rgb}{0.21,0.49,0.74}
\usepackage[pagebackref,breaklinks,colorlinks,allcolors=cvprblue]{hyperref}
\hypersetup{pdfauthor={Guiqiu Liao, Matja\v{z} Jogan, Daniel A. Hashimoto},pdftitle={Compress to Ground: Spatially Addressable Slot Tokens for Multimodal Large Language Models}}

\def\paperID{*****}
\def\confName{CVPR}
\def\confYear{2027}
\title{Compress to Ground: Spatially Addressable Slot Tokens for Multimodal Large Language Models}
\author{Guiqiu Liao$^{1}$ \qquad Matja\v{z} Jogan$^{1}$ \qquad Daniel A. Hashimoto$^{2}$\\
$^{1}$University of Pennsylvania \qquad $^{2}$Duke University\\
{\ttfamily\small guiqiu@engineering.upenn.edu, mjogan@engineering.upenn.edu, daniel.hashimoto@duke.edu}}

\begin{document}
\maketitle
\begin{abstract}
Multimodal large language models (MLLMs) for scene understanding face a joint challenge: driving down the cost of visual inputs through compression while enabling generated answers to reference identifiable image regions. Explicit spatial references make model outputs easier to explain and can facilitate semantic tracking of object interactions, both of which are crucial for surgical scene understanding in robotic surgery. We introduce Spatially Addressable Slot Tokens (\methodname{}), compact token representations of image regions discovered through self-supervised Slot Attention (SA), which serve as both visual inputs and spatial referents for generated answers. Slot tokens are enriched with the appearance of their corresponding regions, and each is associated with an explicit area ID. This allows a grounding curriculum that uses supervision from semantic descriptors, area IDs, and localization to transfer SA latent knowledge and image-specific object-to-area correspondence into the language space, while progressively internalizing such knowledge before inference. We evaluate our method on four surgical datasets covering recognition, localization, and visual question answering. Our method reduces the visual-token count by over 96\% relative to the dense baseline and achieves the highest grounding scores among the evaluated state-of-the-art compression methods, significantly reducing the performance gap compared with dense tokenization. Code is available at \url{https://github.com/PCASOlab/SAS-Token}.
 
\end{abstract}
\begin{figure}[!t]
    \centering
    \includegraphics[width=\columnwidth,pagebox=cropbox]{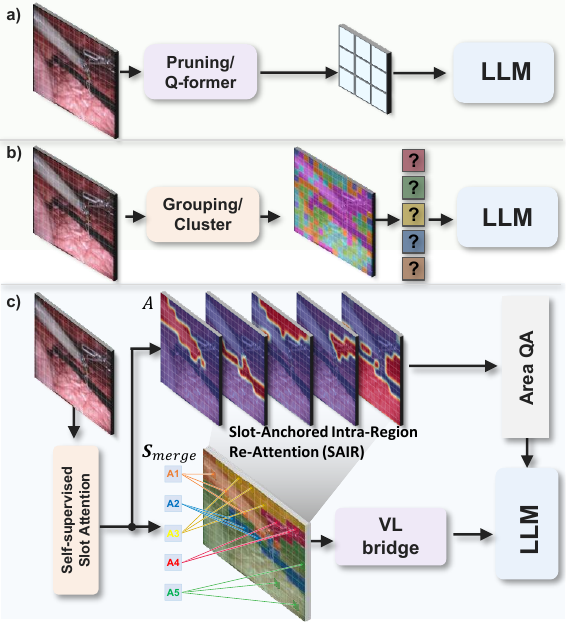}
    \caption{\textbf{Visual token compression methods for MLLMs.} a): A projector maps patch features into language-model tokens. b: Clustering or grouping produces compact tokens without explicit area IDs. c): Our method retains merged slots and their spatial masks, uses SAIR to attend to region features, and pairs the projected visual tokens with area IDs to form \methodname{}. These IDs let the LLM refer to regions represented by its compact visual input.}
    \label{fig:slot2text-overview}
\end{figure}
\section{Introduction}
Multimodal large language models (MLLMs) support recognition, visual question answering (VQA), and localization through a shared language interface~\cite{liu2023visualinstruction} and can thus power AI applications in critical domains such as image-guided surgery~\cite{wang2024surgicallvlm,wang2025endochat}. When interpreting surgical scenes, answers often refer to multiple visually similar instances of surgical instruments and anatomical structures, and to interactions that require precise spatial grounding. For example, identifying a cutting action does not identify which of the instruments is doing the cutting and what part of the anatomy is being cut. MLLMs thus need explicit region references that make this distinction feasible.

While dense visual inputs support grounding through spatial descriptions~\cite{ma2024groma,guo2024regiongpt} or external vision decoders~\cite{lai2023lisa,rasheed2024glamm}, they incur a high token cost. Compressing the visual input introduces a trade-off between the number of visual tokens and grounding accuracy and highlights a need for compact representations that retain useful spatial information. Prior work has established compact visual inputs through simple compression approach like pruning, Q-Former and projection~\cite{shang2024llavaprumerge,li2023blip,li2025tokenpacker} (Fig.~\ref{fig:slot2text-overview}\textbf{a}). Recent grouping, clustering  and segmentation approaches have demonstrated the potential to construct visual tokens that encode semantically coherent image regions ~\cite{li2025tokenpacker,jin2024chatunivi,lee2026divt,lei2026core} (Fig.~\ref{fig:slot2text-overview}\textbf{b}). Self-supervised object discovery offers a principled way to segment the scene using compact, learned region representations and spatial masks, and Slot Attention (SA) has been demonstrated to be one of the most promising object-centric paradigms~\cite{locatello2020object,seitzer2022bridging,didolkar2025transfer}. Prior work in compressing MLLM tokens with slot attention~\cite{chi2025slotmllm,xuslotVLM} was not optimized for spatial grounding due to the permutation invariance of slots. This leads to difficulty in predicting spatial facts: the same instrument can occupy different slots in different images or decompositions. In addition, while slot latents are optimized for grouping, they may discard appearance details needed for recognition and localization.

Motivated by these challenges,  we aim to learn spatially addressable compressed slot tokens through self-supervised object discovery, and use the slot regions (represented as text) to transfer SA knowledge in the latent space to the LLM, enabling direct spatial reasoning. We thus propose to train MLLMs using Spatially Addressable Slot Tokens (\methodname{}) (Fig.~\ref{fig:slot2text-overview}\textbf{c}). First, dynamic slot merging yields region tokens with decoded masks and area IDs. Then, Slot-Anchored Intra-Region Re-Attention (SAIR) generates region-specific appearance representations guided by slot masks. A three-phase curriculum is used to learn object-to-area correspondence. At inference, the LLM can answer surgical questions and can reference corresponding regions while using orders of magnitude fewer tokens compared to models with dense visual mapping.
 Our contributions are threefold. First, we combine dynamic slot merging and SAIR for compact visual tokenization where region tokens serve as the LLM's visual input and provide spatial support for generated area references. Second, we introduce a grounding curriculum that teaches image-specific object-to-area correspondence while removing textual mask guidance, transferring knowledge between SA and the LLM that supports grounded inference. Third, we evaluate across four surgical datasets with multiple vision and language backbones. Our method uses over 96\% fewer visual tokens than the dense baseline and achieves the highest grounding scores among the evaluated state-of-the-art compression methods, significantly narrowing the performance gap compared to dense visual input.
\begin{figure*}[t!]
    \centering
    \includegraphics[width=1.0\textwidth,pagebox=cropbox]{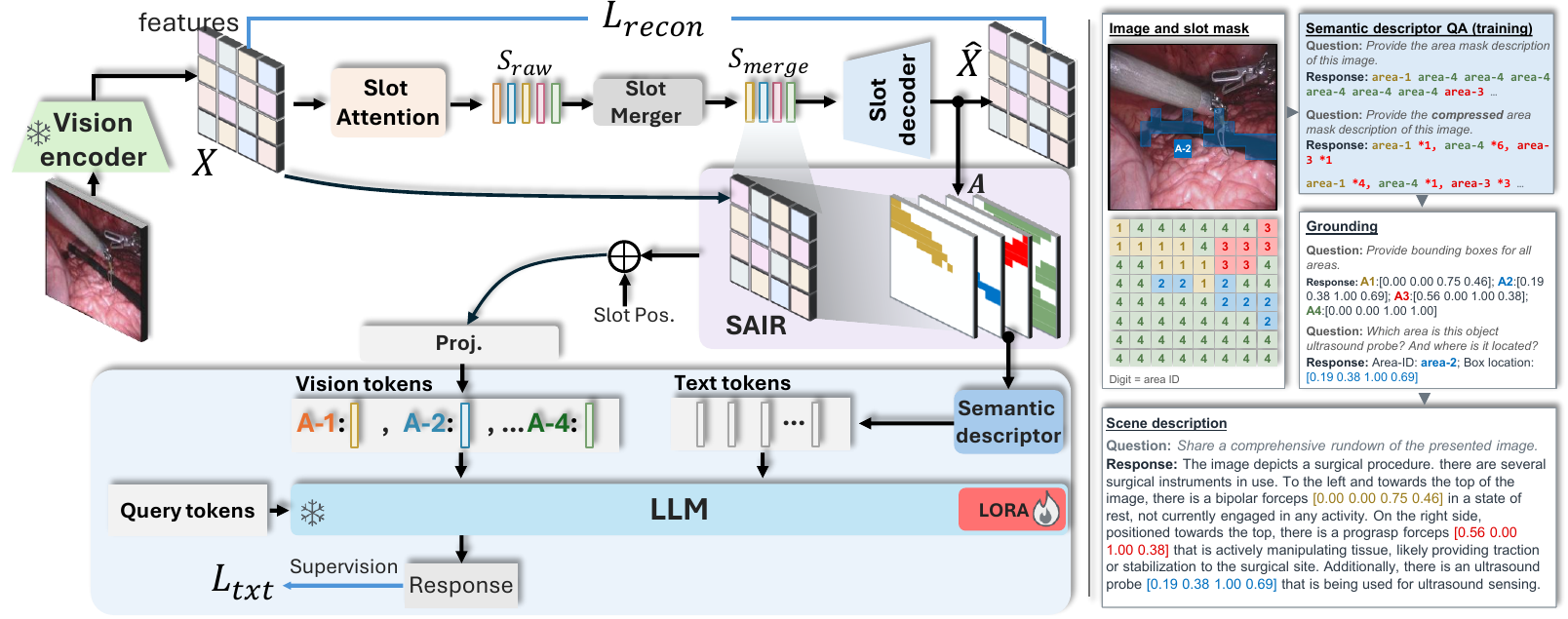}
    \caption{\textbf{Overview of an MLLM pipeline with \methodname{}.} Slot Attention and dynamic merging produce compact region representations with decoded masks. SAIR retrieves appearance features, which are fused with slot and geometric information and projected into area-labeled visual tokens. Right: examples of intermediate mask supervision and grounded responses derived from semantic descriptors. }
    \label{fig:method-training}
\end{figure*}

\section{Related Work}
\noindent\textbf{Multimodal Large Language Models.}
MLLMs integrate pretrained vision encoders with LLMs through learned interfaces, supporting visual question answering and multimodal dialogue~\cite{alayrac2022flamingo,li2023blip,liu2023visualinstruction,dai2023instructblip,chen2023minigptv2}. Subsequent studies improve multimodal pretraining, visual representations, and instruction tuning~\cite{lin2024vila,chen2024internvl,chen2024how,wang2024qwen2vl}. Compact variants explore smaller language backbones and efficient multimodal designs~\cite{chu2024mobilevlmv2,beyer2024paligemma,yao2024minicpm}.

Grounded MLLMs connect language to spatial locations and regions~\cite{peng2023kosmos2,chen2023shikra,you2023ferret}. Groma supports explicit references to detected regions, while RegionGPT encodes supplied regions alongside image-level features~\cite{ma2024groma,guo2024regiongpt}. Surgical-LVLM and EndoChat adapt grounded visual question answering to surgical scenes~\cite{wang2024surgicallvlm,wang2025endochat}. Our work investigates how to retain explicit region references when a compact set of automatically discovered region tokens constitutes the entire visual embedding input.

\noindent\textbf{Efficient Visual Tokenization for MLLMs.}
Visual-token compression uses learned-query summarization~\cite{alayrac2022flamingo,li2023blip,dai2023instructblip}, locality-aware projection~\cite{cha2024honeybee,chu2024mobilevlmv2,li2025tokenpacker}, and spatial downsampling~\cite{chen2024how,yao2024deco}. Earlier vision transformers also investigate dynamic pruning, merging, and clustering~\cite{rao2021dynamicvit,bolya2023tome,zeng2022tcformer}.

Within MLLMs, redundant tokens can be selected or aggregated before entering the language model~\cite{shang2024llavaprumerge,yang2025visionzip,xu2024freepruner,alvar2025divprune}, or reduced within its decoder layers~\cite{chen2024fastv,xing2025conical,ye2025atpllava,lin2025vtw,chen2026v2drop}.

Semantic grouping provides a complementary way to organize visual content~\cite{xu2022groupvit}. Chat-UniVi progressively clusters tokens into multiscale representations for images and videos~\cite{jin2024chatunivi}, while SeTok forms semantic units with an image-dependent token count and supports referring segmentation with an additional segmentation module~\cite{wu2025setok}. DiVT uses similarity-based patch clustering and centroid-query attention restricted to each cluster, supporting adjustable granularity and adaptive token budgets~\cite{lee2026divt}. Our focus is learning an explicit correspondence between generated area identifiers and regions already represented by the compressed input tokens.

\begingroup
\widowpenalty=10000
\noindent\textbf{Object-Centric Vision--Language Models.}
One line of object-centric vision--language modeling uses detected objects or mask-defined regions as visual units~\cite{ma2024groma,yuan2024osprey,zhang2024omgllava}. CORE and AdaTok compress visual inputs by pooling features within masks predicted by pretrained segmentation models~\cite{lei2026core,zhang2025adatok}, while SVP combines SAM-derived regions with geometry embeddings and attention-based feature aggregation~\cite{yang2025repurposing}.
\par\endgroup

Self-supervised object discovery provides a complementary route. Slot Attention supports self-supervised learning of exchangeable object-centric representations through competitive grouping~\cite{locatello2020object}. Recent work has advanced slot-based object discovery in complex real-world images through learning from pretrained visual features~\cite{seitzer2022bridging,kakogeorgiou2024spot,didolkar2025transfer,tian2025pay,bock2026mufasa,fan2024adaptive} and diffusion-based reconstruction~\cite{wu2023slotdiffusion,akan2025slotadapt,nguyen2026coda}. Surgical approaches explore temporal learning of object-centric representations in video~\cite{liao2025slot,liao2025future}.

Slot-VLM connects object and event slots to an LLM for video
understanding~\cite{xuslotVLM}, while Slot-MLLM supports multimodal
understanding and image generation using continuous slot inputs and
quantized visual outputs~\cite{chi2025slotmllm}.
However, object-centric tokenization alone does not establish an
explicit correspondence between generated region references and the
spatial support of the input tokens. For exchangeable slots, an
index has no fixed object identity or spatial meaning~\cite{locatello2020object};
the language model must therefore learn which region each identifier
denotes within the current image. Moreover, compact slot latents may
not retain all appearance cues needed for precise grounding.
Our method learns this image-specific correspondence by pairing
retained slot masks with area identifiers and supervising their
selection through a grounding curriculum. SAIR additionally retrieves
mask-guided appearance features from the visual encoder, which are
fused with slot and geometric information to support recognition
and localization.

\section{Method}
Given a surgical image $I$ and question $q$, a multimodal large language model generates an answer sequence according to $p(y\mid I,q)$. For surgical question answering and spatial grounding, we aim to replace the dense visual prefix with compact region representations $V=\{v_i\}_{i=1}^{K_{\mathrm{eff}}}$, preserving answer-relevant visual information and recoverable spatial support for each representation. 

Figure~\ref{fig:method-training} summarizes our method. Slot Attention groups visual-encoder features into candidate slots, which are dynamically merged before their region masks are decoded. Slot-Anchored Intra-Region Re-Attention (SAIR) retrieves appearance features using the retained slots and masks. These features are fused with slot and geometric information, projected into the language embedding space, and paired with area IDs to form the compact visual prefix. A mask-guided curriculum teaches the language model to associate these image-specific IDs with their spatial support, enabling \emph{spatial addressability}: explicit references to regions represented in the visual input. 

\subsection{Object-Centric Representation Learning}

A frozen encoder $\mathcal{E}$ maps image $I$ to an $h\times w$ feature
grid $X=\mathcal{E}(I)\in\mathbb{R}^{N\times d_x}$, where $N=hw$ and
$d_x$ is the feature dimension. We obtain $K$ candidate slots as
$S^{\mathrm{raw}}=\mathcal{S}(X,S^{(0)})=\{s_i\}_{i=1}^{K}$, where
$\mathcal{S}$ applies a learned adapter followed by three Slot Attention
updates~\cite{locatello2020object}, and $S^{(0)}$ contains the initial
queries. Attention assignments compete across slots and differ from
the decoded masks.

\noindent\textbf{Dynamic Slot Merging.}
A fixed slot budget can allocate multiple slots to similar visual
content. We therefore merge redundant candidate representations
before decoding, using pairwise cosine similarity~\cite{liao2025future}:
\begin{equation}
    c_{ij}=\frac{s_i^\top s_j}
    {\lVert s_i\rVert_2\lVert s_j\rVert_2+\epsilon},
    \quad M_{ij}=\mathbf{1}[c_{ij}>\tau].
\label{eq:slot-merging}
\end{equation}
Here, $\tau$ is the merging threshold, $\epsilon$ ensures numerical
stability, and $M_{ii}=1$ includes each candidate in its own neighborhood.

For each candidate, the merger averages the slot latents in its
threshold-defined neighborhood, assigns the result to that neighborhood's
lowest-index position, and suppresses its higher-index positions.
Retained representatives preserve their original index order without spatial
sorting, yielding $S_{merge}=\{s_i\}_{i=1}^{K_{\mathrm{eff}}}$ with
$K_{\mathrm{eff}}\leq K$. Merging precedes decoding in both tokenizer
pretraining and the downstream language path.

\noindent\textbf{Mask Decoding and Feature Reconstruction.}
For each retained slot $i$ and patch $j$, the decoder predicts a feature
component $\widehat{x}_{ij}$ and mask logit $\ell_{ij}$. We normalize logits
across active slots as $A_{ij}=\operatorname{softmax}_{i}(\ell_{\cdot j})$,
then reconstruct the encoder features $\widehat{X}$ :
\begin{equation}
    \widehat{x}_j=\sum_{i=1}^{K_{\mathrm{eff}}}A_{ij}\widehat{x}_{ij},
    \qquad
    \mathcal{L}_{\mathrm{recon}}
    =\frac{\lVert\widehat{X}-X\rVert_F^2}{Nd_x}.
\end{equation}
This objective learns the decomposition
without object-mask annotations~\cite{seitzer2022bridging} ($L_{\mathrm{recon}}$ in
Figure~\ref{fig:method-training}).

The tokenizer outputs $S_{merge}$ and masks $A$, which guide SAIR retrieval
and define spatial support for area IDs.
During language training, $\mathcal{E}$ remains frozen while Slot Attention
and the slot decoder are optimized.
\subsection{Slot-Anchored Intra-Region Re-Attention}
Retained slot latents summarize grouping features but may discard
appearance details needed for recognition
and grounding. SAIR retrieves additional appearance information
using each slot as a query and its decoded mask as a spatial prior.

\noindent\textbf{Mask-Guided Appearance Retrieval.}
Let $r_j\in[0,1]^2$ denote the normalized coordinates of patch $j$.
We construct queries $q_i\in\mathbb{R}^{d_a}$ and keys
$k_j\in\mathbb{R}^{d_a}$ using learned projection matrices
$W_Q$ and $W_K$:
\begin{equation}
    q_i=W_Q\operatorname{LN}(s_i), \quad
    k_j=W_K[\operatorname{LN}(x_j);r_j].
\end{equation}

The compatibility between slot $i$ and patch $j$ is
$a_{ij}=q_i^\top k_j/\sqrt{d_a}$. We combine this score with the
decoded mask to obtain the attention weights and retrieved features:
\begin{equation}
    \alpha_{ij}=\frac{A_{ij}\exp(a_{ij})}
    {\sum_{\ell=1}^{N}A_{i\ell}\exp(a_{i\ell})},
    \quad p_i=\sum_{j=1}^{N}\alpha_{ij}x_j.
\label{eq:sair-pooling}
\end{equation}
Here, $\alpha_{ij}$ measures the contribution of patch $j$ to
slot $i$, and $p_i$ is the resulting appearance vector.
The mask provides a soft spatial prior, while learned compatibility
reweights appearance information within it.
The original patch features serve as the values.

We initialize $W_Q$ to zero, so the initial attention weights
equal the normalized mask weights. SAIR therefore starts from
mask-weighted mean pooling and learns slot-dependent reweighting
during training.

\noindent\textbf{Visual Token Construction.}
We encode each slot's spatial properties as $g_i=[c_i^y;c_i^x;m_i]$,
where $(c_i^x,c_i^y)$ is the normalized mask centroid and
$m_i=\sum_j A_{ij}$ is the soft mask mass.
The centroid and mass encode region location and extent, respectively.
The fusion module combines this embedding with the slot latent
and retrieved visual features:
\begin{equation}
    v_i=\mathcal{P}\!\left(\mathcal{F}([s_i;p_i;g_i])\right).
\end{equation}
Each retained slot contributes one continuous visual token.

\subsection{Area Identification and Grounded Answering}
\label{sec:area-grounding}
We assign each retained slot an area ID and prepend its text embeddings
to the visual token, e.g., $[e(\texttt{area-1:}),v_1]$, where $e(\cdot)$
denotes the embedding sequence produced by the LLM's native tokenizer
and token-embedding layer. The prefix combines $K_{\mathrm{eff}}$ visual
embeddings with area-ID label tokens.
Each ID links its visual token to a decoded mask, enabling region
references in generated answers.

Area IDs are local to each decomposition: the same instrument may occupy
different slots across images or forward passes, reflecting the slots'
\textbf{permutation symmetry}. The model must therefore associate queried
objects with their current region representations rather than memorize
object-to-index mappings.

We assign each target object the area ID of the retained slot whose mask maximizes intersection-over-union with its region, using the visual prefix's decomposition. Normalized bounding-box coordinates and coarse-location labels supplement area-ID supervision. Since area matching provides an additional spatial cue for box prediction, our model answers grounding queries with area IDs and localization outputs, and surgical QA with text. The visual tokenizer retains masks associated with selected IDs for auxiliary supervision.

\begin{table*}[!t]
\centering
\caption{\textbf{Comparison of visual tokenization methods.} Accuracy measures short-phrase QA; AP50/mIoU measure grounding; BLEU measures visual QA. CP: CoPESD. Best/second-best compression scores are bold/underlined.}
\label{tab:table3}
\footnotesize
\setlength{\tabcolsep}{5.0pt}
\renewcommand{\arraystretch}{1.10}
\fitwidetable{%
\begin{tabular}{@{}l@{\hspace{6pt}}c@{\hspace{8pt}}cccc@{\hspace{8pt}}ccc@{\hspace{8pt}}ccc@{\hspace{8pt}}cccc@{}}
\toprule
\multirow{2}{*}{\textbf{Model}} & \multirow{2}{*}{\textbf{Tokens}} & \multicolumn{4}{c}{\textbf{Accuracy}} & \multicolumn{3}{c}{\textbf{AP50}} & \multicolumn{3}{c}{\textbf{mIoU}} & \multicolumn{4}{c}{\textbf{BLEU}} \\
\cmidrule(lr){3-6}\cmidrule(lr){7-9}\cmidrule(lr){10-12}\cmidrule(lr){13-16}
& & C80 & EV-17 & EV-18 & CP & EV-17 & EV-18 & CP & EV-17 & EV-18 & CP & C80 & EV-17 & EV-18 & CP \\
\midrule
Dense Token (baseline) & 257 & 93.8 & 51.3 & 76.7 & 78.2 & 72.9 & 80.5 & 91.2 & 60.1 & 71.4 & 79.0 & 98.4 & 27.0 & 52.4 & 46.2 \\
\midrule
PruMerge+~\cite{shang2024llavaprumerge} & 199.3 & 82.0 & 44.3 & 56.3 & 77.7 & 20.8 & 23.6 & 57.4 & 25.0 & 28.6 & 52.2 & 94.7 & 22.9 & 47.7 & 48.3 \\
Chat-UniVi~\cite{jin2024chatunivi} & 112 & 90.7 & \underline{47.9} & 63.7 & \underline{77.9} & 18.2 & 33.7 & 63.2 & 23.8 & 36.0 & 55.0 & 97.4 & \underline{26.4} & \textbf{52.6} & 46.6 \\
SeTok~\cite{wu2025setok} & 32.5 & 80.3 & 44.7 & 54.2 & 76.2 & 14.8 & 17.5 & 63.7 & 18.6 & 26.3 & 56.9 & 94.0 & 21.0 & 49.0 & \underline{48.4} \\
DiVT~\cite{lee2026divt} & 20.7 & \textbf{92.3} & 47.2 & 61.0 & \underline{77.9} & 23.7 & 32.6 & 70.4 & 26.6 & 35.3 & 60.3 & \underline{97.7} & 25.5 & 49.4 & \textbf{48.9} \\
Slot-MLLM$^{\dagger}$~\cite{chi2025slotmllm} & 9 & 89.5 & 46.4 & 66.5 & 75.1 & \underline{30.9} & \underline{43.3} & 62.3 & \underline{32.1} & \underline{41.2} & 55.6 & 96.8 & 24.4 & 48.4 & 45.3 \\
Slot-VLM$^{\dagger}$~\cite{xuslotVLM} & 9 & 88.8 & 38.3 & 62.3 & 76.3 & 14.4 & 25.5 & 62.8 & 20.5 & 30.7 & 56.2 & 96.9 & 22.7 & 49.9 & 48.0 \\
DeCo$^{\dagger}$~\cite{yao2024deco} & 9 & 89.1 & 46.2 & \underline{69.9} & 76.8 & 24.2 & 39.9 & \underline{72.8} & 28.3 & 38.9 & \underline{61.1} & 97.4 & 21.9 & 48.9 & 47.8 \\
\midrule
\textbf{\methodname{} (ours)} & 8.8 & \underline{91.3} & \textbf{55.1} & \textbf{71.6} & \textbf{78.7} & \textbf{60.6} & \textbf{68.2} & \textbf{80.7} & \textbf{50.1} & \textbf{56.8} & \textbf{65.4} & \textbf{97.9} & \textbf{26.7} & \underline{51.1} & 47.1 \\
\bottomrule
\end{tabular}%
}
\vspace{2pt}
\parbox{\textwidth}{\raggedright\footnotesize $^{\dagger}$ Fixed visual-token count.}
\end{table*}

\subsection{Slot--Language Grounding Curriculum}
\label{sec:curriculum}
To learn object-to-area correspondence under variable slot
ordering, we progressively remove textual mask guidance.
Across three phases, mask descriptions move from prediction
targets to input context and are then removed, while
area-ID and localization supervision remain. 

\noindent\textbf{(1) Mask-Answer Supervision.}
In this first phase, we supervise mask descriptions alongside surgical QA,
area-ID, bounding-box, and coarse-location responses.
Following the semantic-descriptor approach of
Text4Seg~\cite{lan2025text4seg} (Figure~\ref{fig:method-training}, right), these descriptions associate
spatial positions with image-specific area IDs derived from
the current slot decomposition.
We begin with uncompressed descriptions, then use compressed
descriptions encoding the same assignments in fewer tokens.

\noindent\textbf{(2) Mask Conditioning.}
We move the mask descriptions from supervised
outputs to an input prefix and train the model to predict
QA responses.
As conditioning tokens, the descriptions provide explicit spatial context
for learning object-to-area correspondence without contributing directly
to the response loss. This step also accelerates training.

\noindent\textbf{(3) Direct QA.}
Then we remove mask descriptions from both inputs and targets.
Training retains surgical QA, area-ID QA, and auxiliary
area-location QA, including bounding-box and coarse-location
targets.
The model learns to answer directly from the area-labeled visual prefix
and question, with QA targets retaining spatial supervision.
 
All three phases use the same autoregressive
cross-entropy objective.
For a supervised response $y=(y_1,\ldots,y_T)$,
\begin{equation}
    \mathcal{L}_{\mathrm{txt}}
    = -\frac{1}{T}\sum_{t=1}^{T}
    \log p_{\theta}(y_t \mid C, y_{<t}),
    \label{eq:text-loss}
\end{equation}
where $C$ is the conditioning input, $T$ is the number
of supervised response tokens, and $\theta$ denotes
the trainable parameters.
The loss covers only target response tokens: mask descriptions and QA
responses in Phase~1, and QA responses alone in Phases~2 and~3.

 This progressive removal of textual guidance can be seen as transferring spatial knowledge from SA to the LLM, and is
conceptually related to stepwise CoT
internalization~\cite{deng2024internalizecot} and CODI's
continuous CoT self-distillation~\cite{shen2025codi},
with mask descriptions providing intermediate supervision for slot–language grounding.

\section{Experiments}
\begin{table*}[!t]
\centering
\caption{\textbf{Visual compression across vision and language backbones.} Each pair compares dense inputs with dynamic \methodname{}; dense inputs use no slot encoder (--).}
\label{tab:backbone_comparison}
\footnotesize
\setlength{\tabcolsep}{2pt}
\renewcommand{\arraystretch}{1.10}
\fitwidetable{%
\begin{tabular}{@{}lcccc@{\hspace{8pt}}cccc@{\hspace{8pt}}ccc@{\hspace{8pt}}ccc@{\hspace{8pt}}cccc@{}}
\toprule
\multirow{2}{*}{\textbf{Vision}} & \multirow{2}{*}{\textbf{LLM}} & \multirow{2}{*}{\shortstack{\textbf{Slot}\\\textbf{encoder}}} & \multirow{2}{*}{\textbf{Input}} & \multirow{2}{*}{\textbf{Tokens}} & \multicolumn{4}{c}{\textbf{Accuracy}} & \multicolumn{3}{c}{\textbf{AP50}} & \multicolumn{3}{c}{\textbf{mIoU}} & \multicolumn{4}{c}{\textbf{BLEU}} \\
\cmidrule(lr){6-9}\cmidrule(lr){10-12}\cmidrule(lr){13-15}\cmidrule(lr){16-19}
& & & & & C80 & EV-17 & EV-18 & CP & EV-17 & EV-18 & CP & EV-17 & EV-18 & CP & C80 & EV-17 & EV-18 & CP \\
\midrule
\multirow{2}{*}{SigLIP} & \multirow{2}{*}{\shortstack{TinyLlama\\1.1B}} & -- & Dense & 196 & 90.6 & 44.5 & 67.0 & 76.1 & 24.6 & 33.7 & 67.8 & 28.7 & 36.6 & 58.0 & 97.6 & 23.7 & 52.1 & 47.8 \\
 &  & SigLIP & \textbf{\methodname{}} & 10.1 & 91.9 & 50.0 & 68.4 & 75.5 & 38.6 & 35.2 & 75.2 & 36.5 & 38.2 & 61.7 & 97.7 & 21.2 & 52.0 & 47.5 \\
\midrule
\multirow{2}{*}{CLIP$^{\dagger}$} & \multirow{2}{*}{\shortstack{TinyLlama\\1.1B}} & -- & Dense & 196 & 90.6 & 44.3 & 71.0 & 78.4 & 59.7 & 62.7 & 86.4 & 50.3 & 53.5 & 69.8 & 97.4 & 25.2 & 52.1 & 47.9 \\
 &  & CLIP & \textbf{\methodname{}} & 4.9 & 90.1 & 41.9 & 65.7 & 78.0 & 20.8 & 22.7 & 70.7 & 22.5 & 28.4 & 59.4 & 97.4 & 20.7 & 50.4 & 47.7 \\
\midrule
\multirow{2}{*}{Qwen2.5-VL} & \multirow{2}{*}{\shortstack{Qwen2.5-VL\\7B}} & -- & Dense & 484 & 95.0 & 60.6 & 77.2 & 82.1 & 82.2 & 81.3 & 92.2 & 72.7 & 73.7 & 80.7 & 98.6 & 26.2 & 52.8 & 49.5 \\
 &  & DINOv3 & \textbf{\methodname{}} & 8.8 & 93.8 & 57.6 & 75.7 & 80.9 & 64.4 & 69.4 & 80.6 & 55.0 & 58.3 & 64.5 & 98.2 & 24.9 & 52.4 & 49.4 \\
\midrule
\multirow{2}{*}{DINOv3} & \multirow{2}{*}{\shortstack{Qwen2.5-VL\\7B}} & -- & Dense & 484 & 94.9 & 51.7 & 74.1 & 80.0 & 48.7 & 54.1 & 84.3 & 41.6 & 50.1 & 68.0 & 98.5 & 25.0 & 49.6 & 49.0 \\
 &  & DINOv3 & \textbf{\methodname{}} & 8.8 & 94.3 & 57.2 & 73.8 & 79.7 & 76.3 & 74.8 & 84.7 & 60.7 & 61.9 & 68.5 & 98.4 & 26.9 & 52.1 & 48.9 \\
\bottomrule
\end{tabular}%
}
\vspace{2pt}
\parbox{\textwidth}{\raggedright\footnotesize $^{\dagger}$ Only CLIP uses $\tau=0.95$ because its slots tend to overmerge; all other dynamic configurations use $\tau=0.90$.}
\end{table*}

\subsection{Experimental Setup}

\noindent\textbf{Datasets and Evaluation.}
For self-supervised slot pretraining, we use a corpus of more than 740K surgical frames from the MICCAI 2022 SurgToolLoc challenge~\cite{zia2023surgical} and Thoracic Robotic Surgery~\cite{liao2025slot}. This corpus covers laparoscopic, robotic abdominal, and thoracic surgical scenes. Surgical instruction tuning uses Surg-396K~\cite{wang2025endochat}, which contains 396K image--instruction pairs from Cholec80-VQA, EndoVis-VQLA (derived from the EndoVis 2017 and 2018 challenges), and CoPESD. We follow the original training/test split. Additional dataset details are provided in Appendix~\ref{append:datasets-evaluation}.

We evaluate short-answer recognition using accuracy, visual grounding using AP@0.5 and mIoU, and visual question answering using BLEU~\cite{papineni2002bleu}. Area-ID accuracy measures whether the model identifies the queried object's region within the same slot decomposition. Additional Visual QA and Region-Based QA metrics and results are provided in Appendix~\ref{append:additional-results}.

\noindent\textbf{Baselines.}
Our main comparison evaluates visual tokenization methods for surgical recognition, grounding, and question answering. The dense-token baseline retains the full visual patch sequence and maps it into the language embedding space through an MLP projector.

Compression baselines include PruMerge+~\cite{shang2024llavaprumerge}, Chat-UniVi~\cite{jin2024chatunivi}, SeTok~\cite{wu2025setok}, and DiVT~\cite{lee2026divt}, which cover different token selection and grouping strategies at different compression levels. Their visual-token counts are reported alongside task performance. We set fixed-budget methods (Slot-MLLM~\cite{chi2025slotmllm}, Slot-VLM~\cite{xuslotVLM}, and DeCo~\cite{yao2024deco}) to nine output tokens, closely matching our method's 8.79-token average. Our Slot-MLLM adaptation uses its slot Q-Former for surgical image understanding. Slot-VLM is adapted here and trained on the same surgical datasets.  

Appendix~\ref{append:additional-results} provides additional comparisons with zero-shot general-purpose MLLMs and finetuned surgical MLLMs.

\noindent\textbf{Implementation Details.}
We use TinyLlama-1.1B~\cite{zhang2024tinyllama} or the Qwen2.5-VL-7B decoder~\cite{bai2025qwen25vl} as the language backbone, paired with DINOv3 ViT-B/16~\cite{simeoni2025dinov3}, CLIP~\cite{radford2021learning}, SigLIP~\cite{zhai2023sigmoid}, or the native Qwen2.5-VL vision encoder. The slot-formation encoder can differ from the vision backbone queried by SAIR; for example, DINOv3-derived slots guide the aggregation of native Qwen visual features (Table~\ref{tab:backbone_comparison}). The dynamic object-centric tokenizer starts with $K=17$ slots and uses a default merging threshold of $\tau=0.90$ during training and inference; CLIP uses $\tau=0.95$. We additionally evaluate sampled training thresholds (Table~\ref{tab:threshold_ablation}).

We train with AdamW, cosine learning-rate decay, and a 0.03-epoch warm-up. Up to two auxiliary area-location questions are added per original QA, each weighted by 0.25 relative to the original answer loss. Before curriculum training, we align the VL bridge by updating the slot-feature fusion and projection modules, with the visual encoder, slot tokenizer, and LLM frozen. Then during curriculum training, we optimize Slot Attention, the language model, SAIR attention pooler, and projectors while keeping the vision backbone frozen. Further configuration and training details are provided in Appendix~\ref{append:implementation}.

\begin{figure*}[!t]
    \centering
    \includegraphics[width=0.80\textwidth,pagebox=cropbox]{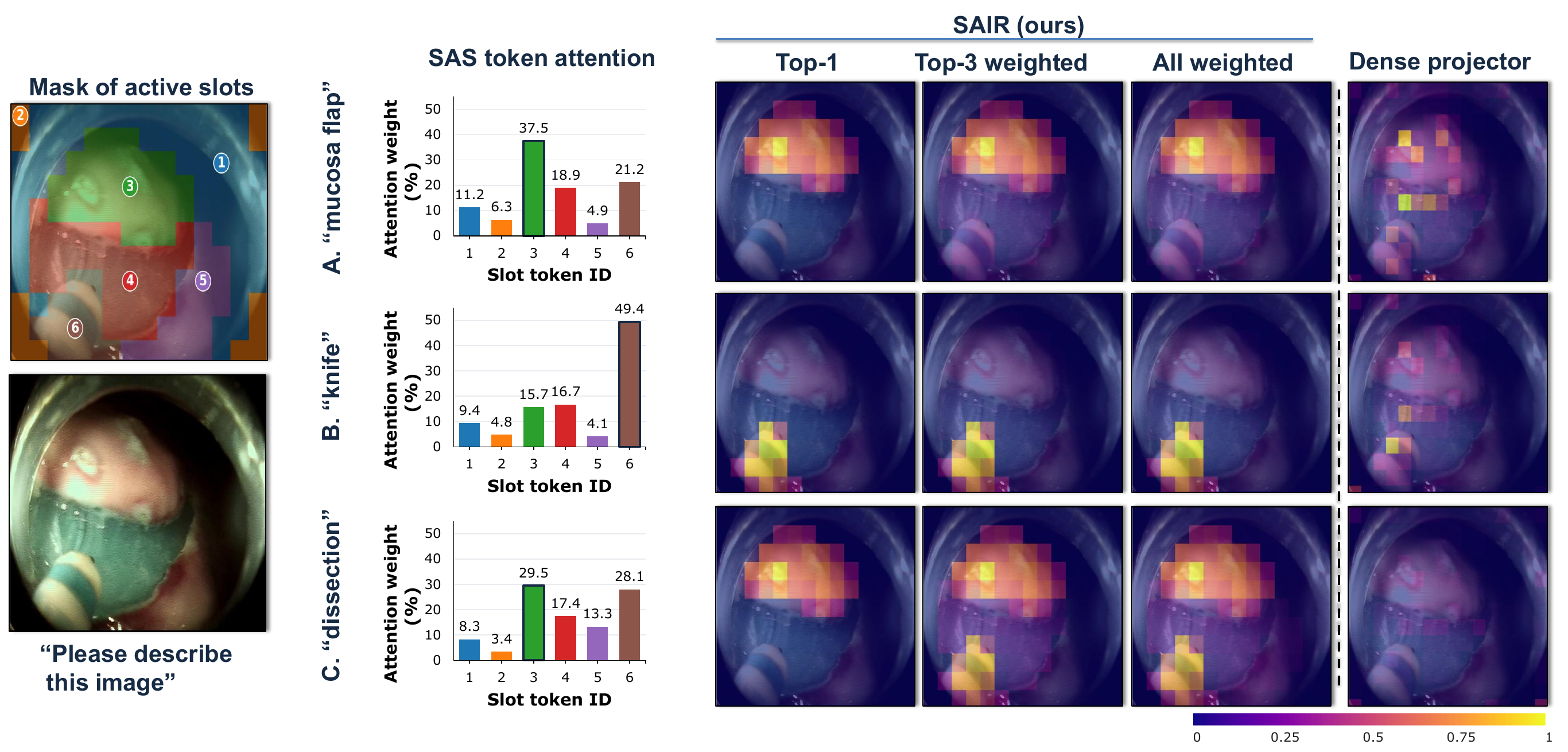}
    \caption{\textbf{Language output-SAS token correspondence under the prompt ``Please describe this image.''} Bars show selected words' attention normalized over active slots. Top-1 shows the most-attended slot's SAIR patch map; Top-3 weighted and All weighted combine maps using word-to-slot weights. Dense-projector maps provide a baseline comparison.}
    \label{fig:slot-language-attention}
\end{figure*}

\subsection{Main Results}

\noindent\textbf{Comparison with Visual Tokenization Methods.}
 We compare SAS tokens against visual tokenization methods by training them using the same language and vision backbone (Table~\ref{tab:table3}). With an average of 8.8 visual tokens, our method reduces the visual embedding count by 96.6\% relative to the 257-token dense baseline. Despite its compact representation, it achieves the highest AP@0.5 and mIoU among the compared compression methods on all three grounding datasets.

The grounding advantage persists at comparable token budgets. Compared with the nine-token Slot-MLLM baseline, our method increases mIoU from 32.1 to 50.1 on EndoVis-17 and from 41.2 to 56.8 on EndoVis-18. On CoPESD, it achieves 65.4 mIoU, exceeding the strongest competing method, DeCo \cite{yao2024deco}, by 4.3 points. The dense baseline nevertheless retains higher grounding scores across all three datasets, indicating a remaining performance cost at this compression level.

Recognition and Visual QA remain competitive, with performance varying by dataset. Among the compression methods, our method achieves the highest short-phrase accuracy on EndoVis-17, EndoVis-18, and CoPESD, and the second-highest on Cholec80. It also obtains the highest Visual QA BLEU on Cholec80 and EndoVis-17, while Chat-UniVi and DiVT lead on EndoVis-18 and CoPESD, respectively. Grounding therefore represents the most consistent advantage of our method in this comparison.

\noindent\textbf{Compression Across Backbones.}
Table~\ref{tab:backbone_comparison} evaluates the \methodname{} implemented with SigLIP, CLIP, and native Qwen2.5-VL vision features, paired with TinyLlama-1.1B or the Qwen2.5-VL-7B  LLM backbone. Across these configurations, dynamic slot representations reduce mean visual-token counts from 196--484 to 4.92--10.09, corresponding to 94.9--98.2\% fewer visual embeddings.

Across the evaluated backbones, our method maintains broadly comparable recognition and Visual QA performance with substantially fewer visual tokens. In particular, with the Qwen2.5-VL-7B decoder, both the native Qwen and DINOv3 vision configurations retain over 95\% of the dense baseline's dataset-averaged recognition accuracy and Visual QA BLEU, using only 8.8 visual tokens instead of 484. The DINOv3/Qwen configuration additionally improves grounding mIoU across all three datasets. With SigLIP and CLIP, recognition and Visual QA scores also remain competitive under substantial compression, and using SigLIP improves grounding mIoU across all three datasets.

\noindent\textbf{Qualitative Slot--Language Correspondence.}
Figure~\ref{fig:slot-language-attention} examines visual correspondence under the prompt ``Please describe this image.'' We combine attention from selected generated words to visual tokens with each slot's fine-grained SAIR attention over image patches. For ``mucosa flap'' and ``knife,'' attention concentrates on the corresponding tissue and instrument slots, respectively. Interestingly, within each region higher attention is obtained at patches with distinctive features (e.g., a lesion on the mucosa), and no language description is provided for these features. For ``dissection,'' attention spans both slots, and the weighted spatial maps highlight both regions. The dense projector produces more spatially dispersed attention in this example, with a less distinct correspondence between the action term and the tissue--instrument combination. These visualizations show the explicit word-to-slot-to-patch correspondence enabled by \methodname{} for generated object and action terms.

\begin{table}[t]
\centering
\caption{\textbf{Comparison of QA supervision and mask-training configurations.} Accuracy: short-phrase QA; EV: average of EndoVis-17/18; CP: CoPESD.}
\label{tab:qa_curriculum_ablation}
\label{tab:mask_supervision_context}
\small
\setlength{\tabcolsep}{5.5pt}
\renewcommand{\arraystretch}{1.10}
\fitcompacttable{%
\begin{tabular}{@{}l@{\hspace{8pt}}ccc@{\hspace{9pt}}cc@{\hspace{9pt}}cc@{}}
\toprule
\multirow{2}{*}{\textbf{Configuration}} & \multicolumn{3}{c}{\textbf{Accuracy}} & \multicolumn{2}{c}{\textbf{AP50}} & \multicolumn{2}{c}{\textbf{mIoU}} \\
\cmidrule(lr){2-4}\cmidrule(lr){5-6}\cmidrule(lr){7-8}
& C80 & EV & CP & EV & CP & EV & CP \\
\midrule
Original QA only & 89.3 & 59.4 & 77.9 & 42.8 & 72.0 & 38.7 & 59.3 \\
Area-ID supervision & 89.3 & 59.1 & 78.1 & 43.2 & 71.7 & 38.3 & 59.2 \\
Mask-prefix conditioning & 89.4 & 58.7 & 77.9 & 42.3 & 74.0 & 37.5 & 60.2 \\
Mask-answer supervision & 89.2 & 59.4 & 78.1 & 38.0 & 75.5 & 34.3 & 60.6 \\
Mask curriculum & 89.5 & 62.5 & 76.8 & 55.1 & 74.5 & 46.5 & 59.0 \\
\textbf{Full training configuration} & 91.3 & 63.3 & 78.7 & 64.4 & 80.7 & 53.5 & 65.4 \\
\bottomrule
\end{tabular}%
}
\end{table}

\begin{table}[t]
\centering
\caption{\textbf{Ablation of visual token construction.} }
\label{tab:pooling_ablation}
\label{tab:component_ablation}
\small
\setlength{\tabcolsep}{5.5pt}
\renewcommand{\arraystretch}{1.10}
\fitcompacttable{%
\begin{tabular}{@{}l@{\hspace{8pt}}ccc@{\hspace{9pt}}cc@{\hspace{9pt}}cc@{}}
\toprule
\multirow{2}{*}{\textbf{Configuration}} & \multicolumn{3}{c}{\textbf{Accuracy}} & \multicolumn{2}{c}{\textbf{AP50}} & \multicolumn{2}{c}{\textbf{mIoU}} \\
\cmidrule(lr){2-4}\cmidrule(lr){5-6}\cmidrule(lr){7-8}
& C80 & EV & CP & EV & CP & EV & CP \\
\midrule
Mask mean only & 89.2 & 59.0 & 77.7 & 37.9 & 63.9 & 37.7 & 55.9 \\
Slots only & 87.6 & 59.1 & 79.8 & 36.1 & 78.4 & 32.0 & 62.8 \\
Mean + fusion & 89.5 & 60.1 & 77.5 & 46.1 & 74.6 & 38.7 & 61.0 \\
\textbf{Full SAIR} & 91.3 & 63.3 & 78.7 & 64.4 & 80.7 & 53.5 & 65.4 \\
\bottomrule
\end{tabular}%
}
\end{table}

\subsection{Further Analysis}
\label{sec:mask-supervision-context}

\begin{figure}[t]
    \centering
    \includegraphics[width=\columnwidth]{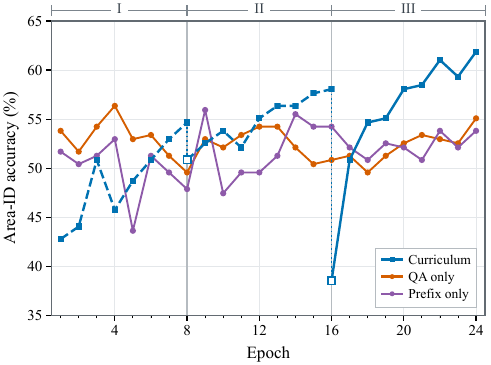}
    \caption{\textbf{Area-ID learning dynamics.} Phases I, II, and III denote mask-answer supervision, mask-prefix conditioning, and QA-only training, respectively. Dashed and solid curriculum segments denote training with and without mask context, respectively. Hollow squares mark immediate context switches.}
    \label{fig:curriculum-endovis17}
\end{figure}

\begin{table}[t]
\centering
\caption{\textbf{Slot-merging threshold sensitivity.} Uniform: randomly samples $\tau_{\mathrm{train}}$ uniformly from $[0.85,0.95]$. }
\label{tab:threshold_ablation}
\small
\setlength{\tabcolsep}{5.5pt}
\renewcommand{\arraystretch}{1.10}
\fitcompacttable{%
\begin{tabular}{@{}ccc@{\hspace{9pt}}cccc@{}}
\toprule
$\boldsymbol{\tau}_{\mathbf{train}}$ & $\boldsymbol{\tau}_{\mathbf{inf}}$ & \textbf{Tokens} & \textbf{Accuracy} & \textbf{AP50} & \textbf{mIoU} & \textbf{BLEU} \\
\midrule
\multirow{3}{*}{\shortstack[l]{ \textbf{0.85} }} & 0.85 & 3.6 & 72.3 & 57.9 & 49.8 & 55.0 \\
 & 0.90 & 6.2 & 73.9 & 65.7 & 54.3 & 54.9 \\
 & 0.95 & 8.8 & 73.4 & 66.7 & 55.2 & \underline{55.5} \\
\midrule
\multirow{3}{*}{\shortstack[l]{ \textbf{0.90} }} & 0.85 & 6.1 & 73.6 & 63.3 & 53.8 & 55.2 \\
 & 0.90 & 8.8 & \underline{74.2} & \textbf{69.8} & 57.4 & \textbf{55.7} \\
 & 0.95 & 11.4 & 74.0 & \underline{69.7} & \textbf{57.9} & 55.4 \\
\midrule
\multirow{3}{*}{\shortstack[l]{ \textbf{0.95} }} & 0.85 & 6.2 & 73.3 & 63.0 & 52.8 & 55.1 \\
 & 0.90 & 8.9 & 74.1 & 68.6 & 56.7 & 55.0 \\
 & 0.95 & 11.7 & \textbf{74.3} & \underline{69.7} & \underline{57.5} & \underline{55.5} \\
\midrule
\multirow{3}{*}{\shortstack[l]{ \textbf{Uniform} }} & 0.85 & 6.0 & 72.9 & 63.2 & 53.3 & 54.7 \\
 & 0.90 & 8.7 & 74.0 & 66.8 & 55.9 & 54.4 \\
 & 0.95 & 11.3 & 73.9 & 67.7 & 56.7 & 54.8 \\
\bottomrule
\end{tabular}%
}
\end{table}

\noindent\textbf{QA Supervision and Mask Curriculum.}
We begin with an original-QA-only baseline that retains the proposed visual-token architecture but excludes additional area-ID supervision, textual mask guidance, and auxiliary area-location QA. This baseline achieves 38.7 EndoVis-average mIoU and 59.3 CoPESD mIoU. Table~\ref{tab:qa_curriculum_ablation} compares different ways of incorporating region supervision: area-ID targets, mask descriptions provided as input context or prediction targets, and a staged mask curriculum. Adding area-ID supervision alone produces little overall change. The mask-prefix and mask-answer variants each extend area-ID supervision and yield dataset-dependent changes, but do not consistently improve grounding. When the mask curriculum is introduced, the performance reaches 46.5 EndoVis-average mIoU, which is a significant improvement. Our full training configuration, which adds synthesized auxiliary area-box QA, achieves the highest recognition and grounding scores across the reported datasets, including 53.5 EndoVis-average mIoU and 65.4 CoPESD mIoU. Both the curriculum and full configurations operate without mask text at evaluation.   These results suggest that progressive removal of mask guidance can transfer region correspondence into the language model, while auxiliary area-box QA further strengthens this spatial grounding.

To further investigate how the mask curriculum develops region correspondence, we track Area-ID accuracy, which measures whether the model identifies the queried object's region token within the current slot decomposition (Figure~\ref{fig:curriculum-endovis17}). Each Area-ID serves as an image-specific discrete address for its corresponding visual token and decoded spatial region, allowing the LLM to refer to the token through language despite the permutation symmetry of Slot Attention. In a focused EndoVis localization setting, all three strategies share the same backbone, model configuration, and pretrained initialization, with 5,285 continuation questions per epoch for 24 epochs. The curriculum progresses through mask-answer supervision, mask-prefix conditioning, and QA-only training, with eight epochs per phase, while the QA-only and prefix-only controls retain their respective settings throughout. The curriculum reaches 61.86\% Area-ID accuracy, exceeding QA-only training (55.08\%) and prefix-only conditioning (53.81\%). For reference, random guessing gives an approximate Area-ID accuracy of 11.36\%. Notably, after mask context is removed in Phase III, Area-ID accuracy recovers from 38.56\% to 61.86\%, indicating that the model can retain and recover image-specific token-to-region correspondence without continued access to explicit mask descriptions. Since Area-ID targets are defined relative to the current slot decomposition, this metric specifically evaluates whether language can address the appropriate visual token rather than serving as an independent localization metric; complementary analyses of slot coverage and localization are provided in Appendix~\ref{append:full-ablations}.

\noindent\textbf{Visual Token Construction.}
We examine how visual-token construction affects recognition and grounding, focusing on the contribution of learned appearance retrieval. All four configurations use the same full training setup and evaluation protocol. They differ in visual-token construction: (1) \emph{Mask mean only} uses mask-weighted mean visual features; (2) \emph{Slots only} uses slot latents; (3) \emph{Mean + fusion} combines mask-weighted mean features with slot latents and geometric information; and (4) \emph{Full SAIR} replaces mean pooling with learned slot-anchored re-attention within the same fusion structure. Table~\ref{tab:pooling_ablation} shows that Full SAIR achieves the highest grounding scores and C80/EndoVis recognition accuracy. Relative to Mean + fusion, Full SAIR improves EndoVis-average mIoU from 38.7 to 53.5 and CoPESD mIoU from 61.0 to 65.4, supporting the benefit of learned appearance retrieval. Slots only achieves the highest CoPESD recognition accuracy (79.8), but lower grounding mIoU of 32.0 on EndoVis and 62.8 on CoPESD. These results suggest that appearance retrieval complements slot representations for spatial grounding.

\noindent\textbf{Slot-Merging Threshold Sensitivity.}
We compare our method trained with three fixed merging thresholds ($0.85$, $0.90$, and $0.95$) or a uniformly sampled threshold (Table~\ref{tab:threshold_ablation}). Each trained model is evaluated at three inference thresholds without changing its weights. A higher inference threshold $\tau_{\mathrm{inf}}$ merges fewer slots, retaining more visual tokens. Across all four training settings, increasing $\tau_{\mathrm{inf}}$ from $0.85$ to $0.95$ improves average grounding mIoU while increasing the average SAS token count. Using $\tau_{\mathrm{inf}}=0.90$ achieves $57.4$ mIoU with $8.8$ tokens; retaining more tokens does not mean improvement across all metrics. These results show that the inference threshold can adjust the token budget and grounding performance of a trained model without retraining.

\section{Conclusion}
We introduced \methodname{}, a compact visual representation that enables surgical MLLMs to explicitly reference image regions. Our method combines dynamic slot merging with SAIR to construct region tokens, while a three-phase curriculum teaches region referencing and localization without textual mask descriptions at inference. More encouragingly, attention links generated object and action terms to relevant SAS tokens and image regions, revealing word-to-slot-to-patch correspondence. Experiments across four surgical datasets show that our method reduces visual-token counts by 96.6\% relative to the dense baseline and achieves the strongest grounding performance among the evaluated compression methods, while maintaining competitive recognition and Visual QA performance.

\iftoggle{cvprfinal}{
\section*{Acknowledgments}
This work was supported by the Linda Pechenik Montague Investigator Award, the American Surgical Association Foundation Fellowship, and the Penn AI Fellowship.
}{}
{
\small 
\bibliographystyle{ieeenat_fullname}
\bibliography{bibliography/ref,table/endochat_references}
}
\clearpage
\appendix
\counterwithin{table}{section}
\counterwithin{figure}{section}
\counterwithin{equation}{section}
\section*{LLM Usage Statement}
We used ChatGPT, Claude, Gemini, and DeepSeek to assist with code development, plotting experimental results, and checking and refining the language of the manuscript. All AI-assisted content was subject to human review, refinement, and verification. The authors take full responsibility for the manuscript, code, analyses, and reported results.

\section{Additional Slot Attention Details}
\label{append:slot_attention}

Slot Attention~\cite{locatello2020object} iteratively groups visual features into a compact set of $K$ slot representations. Following the notation in the main paper, $i$ indexes slots and $j$ indexes image patches. A learned adapter maps the frozen encoder features $X=\mathcal{E}(I)$ to features $Z=\{z_j\}_{j=1}^{N}$ used for slot formation. Our reference configuration uses $K=17$ randomly initialized slots $s_i$ of dimension $d_{\mathrm{SA}}=256$ and three Slot Attention iterations. Dynamic merging subsequently reduces the number of retained slots to $K_{\mathrm{eff}}\leq K$.

At iteration $t$, attention queries are obtained by projecting the layer-normalized previous slot states $\bar{s}_i^{(t-1)}=\operatorname{LN}(s_i^{(t-1)})$, while keys and values are obtained by projecting the layer-normalized adapted patch features. Denoting these projected representations by $q_i^{(t)}$, $k_j$, and $v_j^{\mathrm{SA}}$, respectively, the competitive attention assignments are
\begin{equation}
 B_{ij}^{(t)}
 = \frac{\exp\!\left((q_i^{(t)})^\top k_j/\sqrt{d_{\mathrm{SA}}}\right)}
 {\sum_{r=1}^{K}\exp\!\left((q_r^{(t)})^\top k_j/\sqrt{d_{\mathrm{SA}}}\right)}.
\label{eq:slot-attn-assignment}
\end{equation}
The superscript $\mathrm{SA}$ distinguishes the projected values from the final visual tokens in the main paper. Normalization across slots encourages competition for each input patch. The assignments are then stabilized and normalized across patches to aggregate information for each slot:
\begin{equation}
 \omega_{ij}^{(t)}
 = \frac{B_{ij}^{(t)}+\epsilon}
 {\sum_{\ell=1}^{N}(B_{i\ell}^{(t)}+\epsilon)},
 \qquad
 u_i^{(t)}=\sum_{j=1}^{N}\omega_{ij}^{(t)}v_j^{\mathrm{SA}},
\label{eq:slot-attn-weight}
\end{equation}
where $\epsilon=10^{-8}$. The aggregated feature updates the slot through a gated recurrent unit:
\begin{equation}
 s_i^{(t)}=\operatorname{GRU}\!\left(u_i^{(t)},\bar{s}_i^{(t-1)}\right),
 \qquad t=1,2,3.
\label{eq:slot-attn-update}
\end{equation}
Here, $u_i^{(t)}$ is the GRU input and $\bar{s}_i^{(t-1)}$ is the hidden state supplied to the GRU. The reference configuration disables the optional MLP refinement after the GRU update.

\noindent\textbf{Internal Assignments and Decoded Masks.}
The assignments $B^{(t)}$ are internal to slot formation and are distinct from the decoded masks $A$ in the main paper. After the final Slot Attention iteration, dynamic merging combines redundant slot representations~\cite{liao2025future}. The decoder then produces feature components and spatial masks for the retained slots. These decoded masks support feature reconstruction, guide SAIR appearance retrieval, and provide the spatial regions associated with area IDs. Thus, SAIR uses the masks decoded after merging, rather than the internal Slot Attention assignments.

\noindent\textbf{Slot Identification and Spatial Interpretation.}
The shared Slot Attention updates are permutation-equivariant with respect to the initial slot order: permuting the initial slots correspondingly permutes the updated slots. Slot indices therefore do not encode fixed semantic identities. Moreover, competitive grouping does not guarantee a one-to-one correspondence between slots and semantic objects; an object may span multiple slots, and a slot may include background or multiple structures. Area IDs consequently refer to regions in the current image decomposition, and the grounding curriculum teaches the language model to associate queried objects with those regions.

\section{Additional Details on Datasets and Experimental Protocol}
\label{append:datasets-evaluation}

\noindent\textbf{Slot Pretraining Data: MICCAI 2022 SurgToolLoc.}
The SurgToolLoc challenge~\cite{zia2023surgical} released 24,695 video clips of robot-assisted surgical training exercises recorded on the da Vinci platform. Each clip is 30 seconds long, captured at 60~FPS, and accompanied only by clip-level tool-presence labels, which we discard. This source offers wide variation in instrument type, tissue appearance, camera viewpoint, and instrument--tissue interaction in robotic abdominal scenes.

\noindent\textbf{Slot Pretraining Data: Thoracic Robotic Surgery.}
The thoracic dataset~\cite{liao2025slot} contains 264 clips of 30 frames each, collected from 40 robot-assisted right upper lobectomy procedures. Although much smaller than SurgToolLoc, it comes from real clinical cases rather than training exercises and covers a different anatomical region, with lung parenchyma, hilar vessels, and bronchial structures, as well as bleeding and smoke that are largely absent from SurgToolLoc.
 
\noindent\textbf{Surgical Instruction-Tuning Data.}
For surgical visual instruction tuning, we build on the Surg-396K dataset~\cite{wang2025endochat} and follow the same official training/test split. Surg-396K contains approximately 41.4K endoscopic images and 396K image--instruction pairs collected or generated from three constituent datasets: Cholec80-VQA, EndoVis-VQLA, and CoPESD. Cholec80-VQA contributes 21,591 frames sampled from 40 laparoscopic cholecystectomy videos, together with classification-style and sentence-level question--answer pairs. EndoVis-VQLA contributes 2,104 frames from the EndoVis 2017 and 2018 robotic-surgery datasets, with question--answer annotations and object-level bounding boxes. CoPESD contributes 17,679 images extracted from more than 35 hours of endoscopic submucosal dissection videos, with annotations describing surgical instruments, target tissues, motions, motion directions, and their corresponding spatial regions. These datasets provide complementary tests of the
  proposed compact visual tokenization method.

\noindent\textbf{Area-ID Targets and Evaluation.}\label{append:area-targets}
For a target region $T$ and retained-slot hard masks $A_i^{\mathrm{h}}$, the reference area ID is
\begin{equation}
 i^\star=\arg\max_{1\leq i\leq K_{\mathrm{eff}}}
 \operatorname{IoU}(A_i^{\mathrm{h}},T).
\label{eq:area-id-target}
\end{equation}
Targets use the visual prefix's current decomposition and are recomputed when it changes. Area-ID accuracy measures exact identifier matches on eligible grounding queries; invalid IDs count as incorrect. IDs are image-specific, so baselines without the same area vocabulary are compared only through recognition, localization, and VQA metrics.

 
\section{Additional Configuration Details}
\label{append:implementation}
The DINOv3/TinyLlama reference uses frozen DINOv3 features for slot formation and SAIR, paired with TinyLlama-1.1B. For this configuration, inputs are $224\times224$ RGB images, yielding 196 patches with 768-dimensional features; SAIR uses 128-dimensional attention. The merging threshold is 0.9, with slot sorting disabled, and $K_{\mathrm{eff}}$ is measured after merging. Evaluation uses greedy decoding with seed 0. The Qwen2.5-VL backbone comparisons use 484 dense visual tokens before SAS-token compression. In the native-Qwen dynamic configuration, DINOv3-derived slots guide aggregation of native Qwen visual features, and evaluation uses no mask-text prefix.

After VL-bridge alignment, the curriculum comprises three phases: mask-answer supervision using uncompressed and then compressed descriptions; conditioning on compressed mask prefixes; and QA-only training that retains original surgical QA, area-ID QA, and auxiliary area-location QA.

For each original QA pair, auxiliary area-location QA sampling attempts to select one area with an unambiguous ground-truth (GT) match and one unmatched occupied area, falling back to other eligible, distinct areas as needed. Empty areas and areas with conflicting GT matches are excluded. Each auxiliary area-location answer loss has weight 0.25, and the original answer loss has weight 1. The weighted sum is normalized by $1.5$ times the number of original QA pairs, even when fewer auxiliary questions are eligible.

\section{Additional Results}
\label{append:additional-results}

\begin{table*}[!htbp]
\centering
\caption{\textbf{Comparison with specialized surgical models and zero-shot general-purpose and biomedical MLLMs.} Acc and F-score evaluate Single Phrase QA. Visual and Region report BLEU on sentence-level answers for Visual QA and Region-Based QA, respectively, following the task definitions of Surg-396K~\cite{wang2025endochat}.}
\label{tab:qa_comparison_merged_appendix}
\small
\setlength{\tabcolsep}{4.0pt}
\renewcommand{\arraystretch}{1.08}
\begin{tabular*}{\textwidth}{@{\extracolsep{\fill}}l*{9}{c}@{}}
\toprule
\multirow{2}{*}{Model} & \multicolumn{3}{c}{Cholec80} & \multicolumn{2}{c}{EndoVis-18} & \multicolumn{2}{c}{EndoVis-17} & \multicolumn{2}{c}{CoPESD} \\
\cmidrule(lr){2-4}\cmidrule(lr){5-6}\cmidrule(lr){7-8}\cmidrule(lr){9-10}
 & Acc & F-score & Visual & Region & Visual & Region & Visual & Region & Visual \\
\midrule
MedFuse~\cite{sharma2021medfusenet} & 86.1 & 30.9 & 33.3 & -- & -- & -- & -- & -- & -- \\
VisualBERT~\cite{li2019visualbert} & 89.7 & 63.3 & 95.6 & -- & -- & -- & -- & -- & -- \\
VisualBERT ResMLP~\cite{seenivasan2022surgicalvqa} & 89.8 & 63.4 & 95.2 & -- & -- & -- & -- & -- & -- \\
Surgical-LVLM~\cite{wang2024surgicallvlm} & 87.5 & 60.1 & 95.1 & -- & -- & -- & -- & -- & -- \\
BiomedGPT~\cite{zhang2023biomedgpt} & -- & -- & -- & 2.2 & 6.6 & 3.6 & 8.8 & 1.7 & 1.6 \\
LLaVA-Med~\cite{li2024llavamed} & -- & -- & -- & 4.7 & 13.5 & 9.0 & 12.9 & 6.7 & 4.6 \\
SPHINX~\cite{lin2023sphinx} & -- & -- & -- & 2.6 & 15.1 & 3.3 & 14.1 & 6.2 & 7.0 \\
EndoChat~\cite{wang2025endochat} & 92.1 & 61.6 & 96.8 & 59.7 & 52.2 & 18.1 & 21.8 & 49.8 & 46.9 \\
\methodname{} (final) & 91.3 & 69.5 & 97.9 & 61.0 & 51.1 & 19.9 & 26.7 & 50.9 & 47.1 \\
\bottomrule
\end{tabular*}
\end{table*}

\noindent\textbf{Comparison with Surgical, General-Purpose, and Biomedical Models.}\label{append:full-specialized-comparison}\label{append:surg396k-split-results}
The main experiments compare visual tokenization methods under shared vision and language backbones. Table~\ref{tab:qa_comparison_merged_appendix} broadens the evaluation to published surgical QA systems and zero-shot general-purpose and biomedical MLLMs, assessing how the compact slot representation compares with established systems on surgical question answering.

Our model and EndoChat~\cite{wang2025endochat} use Surg-396K for surgical instruction tuning. The specialized surgical QA baselines use dataset-specific training settings~\cite{wang2024surgicallvlm,sharma2021medfusenet,li2019visualbert,seenivasan2022surgicalvqa}. while the other MLLMs ~\cite{lin2023sphinx,li2024llavamed,zhang2023biomedgpt} are evaluated zero-shot on the surgical benchmarks. Our model additionally uses surgical frames for self-supervised slot pretraining and introduces area-based curriculum supervision. 

The specialized models in the first four rows report results only on Cholec80, and the general-purpose and biomedical MLLMs in the next three rows are evaluated zero-shot only on the Surg-396K splits; a dash marks an unreported result. EndoChat is the most competitive baseline, but it relies on dense visual input, averaging 1,359 language-model tokens per sample in our measurements. The final model remains comparable to EndoChat across datasets, with higher Region-Based QA BLEU on all three Surg-396K splits.

\begin{table*}[!htbp]
\centering
\caption{\textbf{Training and inference cost across slot-merging thresholds with DINOv3 or native Qwen vision and Qwen2.5-VL-7B.} Gray values in parentheses give each cost as a percentage of the dense baseline with the same vision setup. Lower is better.}
\label{tab:qwen_threshold_efficiency}
\footnotesize
\setlength{\tabcolsep}{4pt}
\renewcommand{\arraystretch}{1.2}
\begin{tabular*}{\textwidth}{@{\extracolsep{\fill}}l*{8}{c}@{}}
\toprule
\multirow{2}{*}{Method} & \multicolumn{2}{c}{\textbf{Training} (ms/step)} & \multicolumn{2}{c}{\textbf{Inference} (ms/sample)} & \multicolumn{2}{c}{\textbf{Visual KV} (MiB)} & \multicolumn{2}{c}{\textbf{Prefill} (ms)} \\
\cmidrule(lr){2-3}\cmidrule(lr){4-5}\cmidrule(lr){6-7}\cmidrule(lr){8-9}
 & DINOv3 & Native Qwen & DINOv3 & Native Qwen & DINOv3 & Native Qwen & DINOv3 & Native Qwen \\
\midrule
Dense & \begin{tabular}[c]{@{}c@{}}381.50\\\scriptsize\textcolor{gray}{(100.0\%)}\end{tabular} & \begin{tabular}[c]{@{}c@{}}457.39\\\scriptsize\textcolor{gray}{(100.0\%)}\end{tabular} & \begin{tabular}[c]{@{}c@{}}152.20\\\scriptsize\textcolor{gray}{(100.0\%)}\end{tabular} & \begin{tabular}[c]{@{}c@{}}232.22\\\scriptsize\textcolor{gray}{(100.0\%)}\end{tabular} & \begin{tabular}[c]{@{}c@{}}26.47\\\scriptsize\textcolor{gray}{(100.0\%)}\end{tabular} & \begin{tabular}[c]{@{}c@{}}26.47\\\scriptsize\textcolor{gray}{(100.0\%)}\end{tabular} & \begin{tabular}[c]{@{}c@{}}98.06\\\scriptsize\textcolor{gray}{(100.0\%)}\end{tabular} & \begin{tabular}[c]{@{}c@{}}177.15\\\scriptsize\textcolor{gray}{(100.0\%)}\end{tabular} \\
\addlinespace[3pt]
$\mathrm{SAS\ Token}_{\tau=0.9}$ & \begin{tabular}[c]{@{}c@{}}191.85\\\scriptsize\textcolor{gray}{(50.3\%)}\end{tabular} & \begin{tabular}[c]{@{}c@{}}267.38\\\scriptsize\textcolor{gray}{(58.5\%)}\end{tabular} & \begin{tabular}[c]{@{}c@{}}101.42\\\scriptsize\textcolor{gray}{(66.6\%)}\end{tabular} & \begin{tabular}[c]{@{}c@{}}175.17\\\scriptsize\textcolor{gray}{(75.4\%)}\end{tabular} & \begin{tabular}[c]{@{}c@{}}0.62\\\scriptsize\textcolor{gray}{(2.3\%)}\end{tabular} & \begin{tabular}[c]{@{}c@{}}0.62\\\scriptsize\textcolor{gray}{(2.3\%)}\end{tabular} & \begin{tabular}[c]{@{}c@{}}57.93\\\scriptsize\textcolor{gray}{(59.1\%)}\end{tabular} & \begin{tabular}[c]{@{}c@{}}128.55\\\scriptsize\textcolor{gray}{(72.6\%)}\end{tabular} \\
\addlinespace[3pt]
$\mathrm{SAS\ Token}_{\tau=0.8}$ & \begin{tabular}[c]{@{}c@{}}191.24\\\scriptsize\textcolor{gray}{(50.1\%)}\end{tabular} & \begin{tabular}[c]{@{}c@{}}262.78\\\scriptsize\textcolor{gray}{(57.5\%)}\end{tabular} & \begin{tabular}[c]{@{}c@{}}98.28\\\scriptsize\textcolor{gray}{(64.6\%)}\end{tabular} & \begin{tabular}[c]{@{}c@{}}172.17\\\scriptsize\textcolor{gray}{(74.1\%)}\end{tabular} & \begin{tabular}[c]{@{}c@{}}0.30\\\scriptsize\textcolor{gray}{(1.1\%)}\end{tabular} & \begin{tabular}[c]{@{}c@{}}0.30\\\scriptsize\textcolor{gray}{(1.1\%)}\end{tabular} & \begin{tabular}[c]{@{}c@{}}56.97\\\scriptsize\textcolor{gray}{(58.1\%)}\end{tabular} & \begin{tabular}[c]{@{}c@{}}128.76\\\scriptsize\textcolor{gray}{(72.7\%)}\end{tabular} \\
\addlinespace[3pt]
$\mathrm{SAS\ Token}_{\tau=0.7}$ & \begin{tabular}[c]{@{}c@{}}191.59\\\scriptsize\textcolor{gray}{(50.2\%)}\end{tabular} & \begin{tabular}[c]{@{}c@{}}263.45\\\scriptsize\textcolor{gray}{(57.6\%)}\end{tabular} & \begin{tabular}[c]{@{}c@{}}95.36\\\scriptsize\textcolor{gray}{(62.7\%)}\end{tabular} & \begin{tabular}[c]{@{}c@{}}171.27\\\scriptsize\textcolor{gray}{(73.8\%)}\end{tabular} & \begin{tabular}[c]{@{}c@{}}0.07\\\scriptsize\textcolor{gray}{(0.3\%)}\end{tabular} & \begin{tabular}[c]{@{}c@{}}0.07\\\scriptsize\textcolor{gray}{(0.3\%)}\end{tabular} & \begin{tabular}[c]{@{}c@{}}55.63\\\scriptsize\textcolor{gray}{(56.7\%)}\end{tabular} & \begin{tabular}[c]{@{}c@{}}127.99\\\scriptsize\textcolor{gray}{(72.2\%)}\end{tabular} \\
\addlinespace[3pt]
$\mathrm{SAS\ Token}_{\tau=0.6}$ & \begin{tabular}[c]{@{}c@{}}191.10\\\scriptsize\textcolor{gray}{(50.1\%)}\end{tabular} & \begin{tabular}[c]{@{}c@{}}265.25\\\scriptsize\textcolor{gray}{(58.0\%)}\end{tabular} & \begin{tabular}[c]{@{}c@{}}93.79\\\scriptsize\textcolor{gray}{(61.6\%)}\end{tabular} & \begin{tabular}[c]{@{}c@{}}169.97\\\scriptsize\textcolor{gray}{(73.2\%)}\end{tabular} & \begin{tabular}[c]{@{}c@{}}0.06\\\scriptsize\textcolor{gray}{(0.2\%)}\end{tabular} & \begin{tabular}[c]{@{}c@{}}0.06\\\scriptsize\textcolor{gray}{(0.2\%)}\end{tabular} & \begin{tabular}[c]{@{}c@{}}55.25\\\scriptsize\textcolor{gray}{(56.3\%)}\end{tabular} & \begin{tabular}[c]{@{}c@{}}126.40\\\scriptsize\textcolor{gray}{(71.4\%)}\end{tabular} \\
\bottomrule
\end{tabular*}
\end{table*}

\noindent\textbf{Training and Inference Efficiency.}\label{append:qwen-threshold-efficiency}
Table~\ref{tab:qwen_threshold_efficiency} evaluates whether compact visual tokens translate into practical training and inference savings with Qwen2.5-VL-7B. We compare separately trained dense and \methodname{} checkpoints using either DINOv3 or native Qwen visual features. Within each family, the slot checkpoint is evaluated at four inference thresholds, $\tau_{\mathrm{inf}}\in\{0.60,0.70,0.80,0.90\}$, with unchanged weights. The native-Qwen slot model additionally uses DINOv3 to form its slot anchors. Gray parenthetical values express costs as percentages of the corresponding dense baseline within the same vision setup.

All measurements are performed sequentially on the same RTX 6000 Ada Generation GPU with a batch size of 1. Images are decoded into CPU memory before timing. Timings include fresh preprocessing and visual computation without cached features or slots, and exclude model loading, disk reads, scoring, and report writing. Inference latency is averaged over the same 472 evaluation questions. No mask-text prefix is supplied; slot models retain their area-plus-box prompts and inline area labels. Prefill latency includes preprocessing, visual encoding and fusion, slot extraction where applicable, and the model's forward pass through the vocabulary head. Training-step timing includes preprocessing, forward and backward passes, and the optimizer update.
Across the four thresholds, visual KV-cache memory falls to 0.2--2.3\% of that of the corresponding dense baseline. This measures the cache contribution of visual tokens, not total GPU memory. Training-step time is approximately 50\% of that of the dense baseline with DINOv3 and 58\% with native Qwen vision. Short-answer inference latency is 62--67\% and 73--75\% of that of the respective dense baselines, while prefill latency is 56--59\% and 71--73\%, respectively. The savings therefore extend beyond visual-token storage to measured training and inference time.

\begin{table*}[!htbp]
\centering
\caption{\textbf{Non-surgical image question answering and approximate computational cost.} Benchmark scores are percentages. ScienceQA reports image-subset accuracy; POPE reports mean F1 across three evaluation categories. Visual-token counts refer to image embeddings.}
\label{tab:natural_vqa_comparison}
\footnotesize
\setlength{\tabcolsep}{3pt}
\renewcommand{\arraystretch}{1.2}
\begin{tabular*}{\textwidth}{@{\extracolsep{\fill}}l*{8}{c}@{}}
\toprule
Model & GQA $\uparrow$ & VizWiz $\uparrow$ & \begin{tabular}[c]{@{}c@{}}ScienceQA  $\uparrow$\end{tabular} & POPE $\uparrow$ & \begin{tabular}[c]{@{}c@{}}Visual\\tokens $\downarrow$\end{tabular} & \begin{tabular}[c]{@{}c@{}}Prefill\\(ms) $\downarrow$\end{tabular} & \begin{tabular}[c]{@{}c@{}}Prefix KV\\(MiB) $\downarrow$\end{tabular} & \begin{tabular}[c]{@{}c@{}}Training time\\(ms/step) $\downarrow$\end{tabular} \\
\midrule
LLaVA-1.5-7B~\cite{liu2024llava15} & 62.0 & 50.0 & 66.8 & 85.9 & 576 & 94.70 & 288.000 & 423.42 \\
\methodname{} (1.1B) & 48.73 & 47.35 & 66.34 & 70.30 & 9.13 & 21.61 & 1.054 & 83.11 \\
\methodname{} (7B) & 61.73 & 66.77 & 87.36 & 84.67 & 8.13 & 77.32 & 0.554 & 200.02 \\
\bottomrule
\end{tabular*}
\end{table*}

\noindent\textbf{Non-Surgical Image Question Answering.}\label{append:natural-vqa}
Table~\ref{tab:natural_vqa_comparison} extends the evaluation beyond surgical images to GQA, VizWiz, ScienceQA, and POPE using two \methodname{} model variants: a 1.1B variant based on TinyLlama and a 7B variant based on Qwen2.5-VL. Both variants use the same non-surgical adaptation pool, with 17 initial slots and a merging threshold of 0.9. The pool contains 1,381,205 QA pairs derived from 329,162 source conversation or question records. The number of source records corresponds to 49.5\% of LLaVA-1.5's 665,298 instruction-tuning records.

Using this adaptation pool, which contains fewer source records than LLaVA's instruction-tuning dataset, the 1.1B model (based on TinyLlama) achieves 47.35 on VizWiz and 66.34 on the image subset of ScienceQA, compared with published LLaVA-1.5-7B scores of 50.0 and 66.8. The 7B SAS variant (based on Qwen2.5-VL) achieves 61.73 on GQA, 66.77 on VizWiz, and 87.36 on the image subset of ScienceQA. POPE F1, averaged across the three evaluation categories, is 70.30 for the 1.1B variant and 84.67 for the 7B variant, versus 85.9 for LLaVA-1.5-7B.

The efficiency evaluation uses the same 32 non-surgical image questions for each model, with three inference repeats and 32 separate training updates per model. The 1.1B and 7B \methodname{} variants use an average of 9.13 and 8.13 visual slot tokens, respectively. Their prefill latencies are 21.61 and 77.32\,ms, respectively, compared with 94.70\,ms for the LLaVA timing proxy. The corresponding times for a forward pass, a backward pass, and an SGD update are 83.11, 200.02, and 423.42\,ms per step. The 7B variant's training measurement updates the LoRA parameters and the visual bridge, whereas the LLaVA proxy uses full-model SGD. The LLaVA cost estimates come from a local Text4Seg/LLaVA-7B checkpoint. These estimates and the retained 1.1B measurements were obtained on a different machine from the 7B measurements, but with the same GPU model. These approximate cross-system costs do not isolate the effect of slot tokenization.

\begin{figure*}[t]
    \centering
    \includegraphics[width=0.8\textwidth]{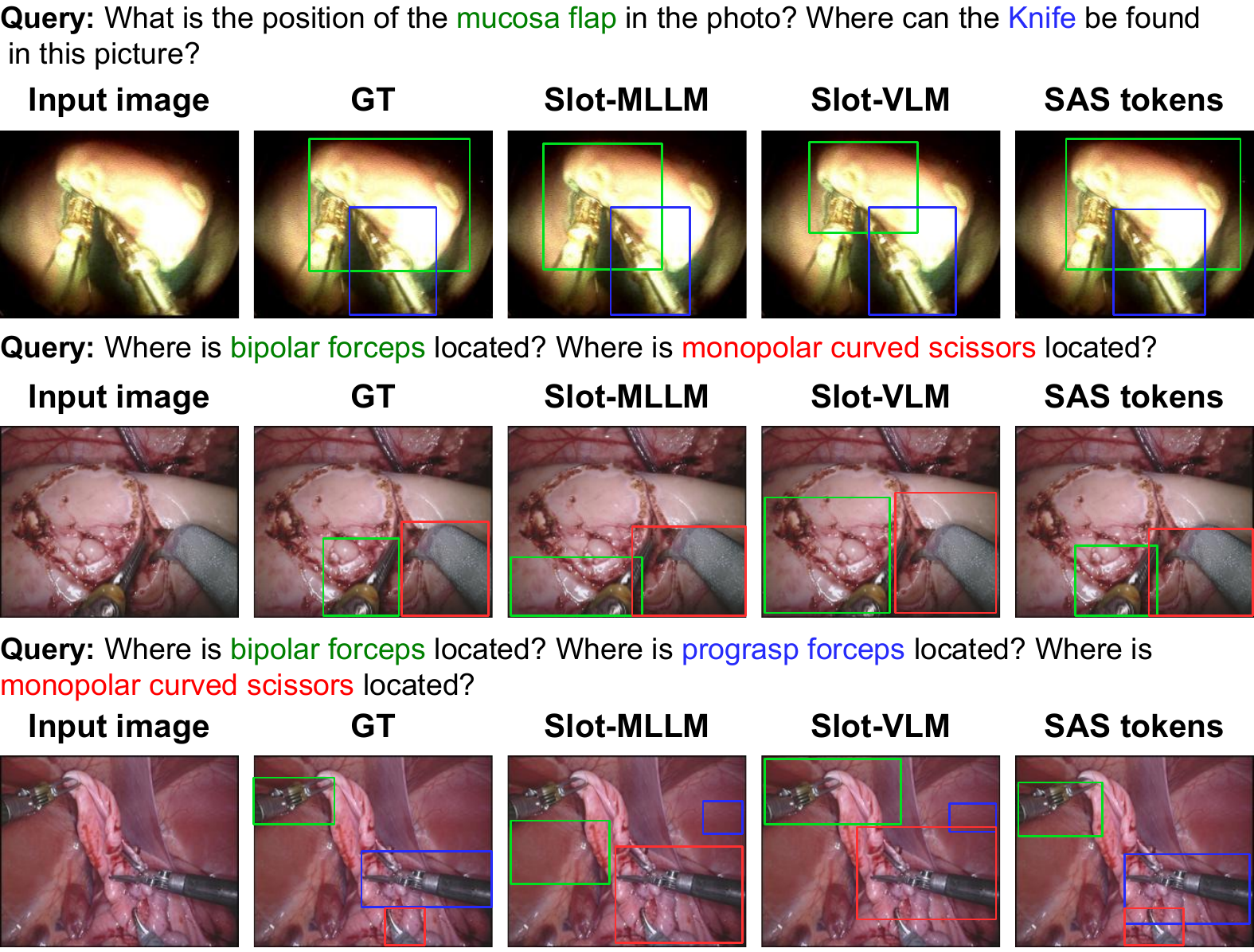}
    \caption{\textbf{Qualitative grounding examples.} Colored boxes indicate the regions corresponding to the color-matched query terms. GT denotes ground-truth boxes.}
    \label{fig:qualitative-grounding}
\end{figure*}

\noindent\textbf{Qualitative Grounding Examples.}\label{append:full-grounding}
Slot-MLLM and Slot-VLM achieve competitive Visual QA BLEU scores of 96.8 and 96.9, respectively, on Cholec80 (Table~\ref{tab:table3}), but these scores do not necessarily imply precise spatial grounding. Figure~\ref{fig:qualitative-grounding} compares their predicted boxes with those of our model. In the two multi-instrument EndoVis examples, the Slot-VLM and Slot-MLLM baselines produce oversized or displaced boxes for the forceps, whereas our predictions align more closely with the ground truth. These examples illustrate the difficulty of preserving precise localization with compact visual representations.

\section{Additional Analysis}
\label{append:full-ablations}

\begin{table*}[!t]
\centering
\caption{\textbf{Area identification and grounding across QA supervision configurations.} ID Acc.: exact match with the reference area ID. Area AP50: IoU $\geq0.5$ for the area-derived box, i.e., the box enclosing the masks of the predicted areas. Direct AP50: IoU $\geq0.5$ for the box generated in the answer. A dash marks a metric that is unavailable (Slot-MLLM) or not applicable (Original QA only). Best scores are bold.}
\label{tab:qa_area_grounding}
\footnotesize
\setlength{\tabcolsep}{2pt}
\renewcommand{\arraystretch}{1.12}
\begin{tabular*}{\textwidth}{@{\extracolsep{\fill}}l*{9}{c}@{}}
\toprule
\multirow{2}{*}{\textbf{Method}} & \multicolumn{3}{c}{\textbf{EV-17}} & \multicolumn{3}{c}{\textbf{EV-18}} & \multicolumn{3}{c}{\textbf{CoPESD}} \\
\cmidrule(lr){2-4}\cmidrule(lr){5-7}\cmidrule(lr){8-10}
& ID Acc. & Area AP50 & Direct AP50 & ID Acc. & Area AP50 & Direct AP50 & ID Acc. & Area AP50 & Direct AP50 \\
\midrule
Slot-MLLM~\cite{chi2025slotmllm} & -- & -- & 30.9 & -- & -- & 43.3 & -- & -- & 62.3 \\
Original QA only & -- & -- & 44.1 & -- & -- & 41.5 & -- & -- & 72.0 \\
Area-ID supervision & 51.7 & 36.4 & 46.6 & 40.2 & 26.0 & 39.9 & 70.3 & 42.4 & 71.7 \\
Mask-prefix conditioning & 47.0 & 33.1 & 42.8 & 40.7 & 26.1 & 41.8 & 71.7 & 42.5 & 74.0 \\
Mask-answer supervision & 47.5 & 32.6 & 41.1 & 36.1 & 23.1 & 34.8 & 72.9 & 43.3 & 75.5 \\
Mask curriculum & 65.3 & 41.5 & 50.8 & 60.8 & 33.9 & 59.3 & 72.5 & 44.3 & 74.5 \\
\textbf{Full training configuration} & \textbf{76.7} & \textbf{58.9} & \textbf{60.6} & \textbf{69.2} & \textbf{48.6} & \textbf{68.2} & \textbf{79.2} & \textbf{52.3} & \textbf{80.7} \\
\bottomrule
\end{tabular*}
\end{table*}

\noindent\textbf{QA Supervision, Area Identification, and Grounding.}
Table~\ref{tab:qa_area_grounding} complements the main-paper QA-supervision ablation (Table~\ref{tab:qa_curriculum_ablation}) by reporting area identification and grounding for the same six checkpoints, alongside the Slot-MLLM baseline. ID Acc. measures exact agreement with the annotation-derived reference area IDs for each checkpoint's decomposition. Area AP50 evaluates the tight bounding box enclosing the predicted slot-mask union. The combination of these two reflects how LLM \textit{identify latent slots} and \textit{express the spatial knowledge} transferred from object-centric learning. 
 Direct AP50 evaluates the box coordinates generated in the answer, which is supervised with the GT coordinates text. Both report the percentage of predictions with continuous-coordinate IoU $\geq0.5$. 

Scores are percentages, and bold values mark the best displayed score in each column. Dashes indicate area-based metrics that are unavailable for Slot-MLLM or not applicable to ``Original QA only'' approach, which does not predict area IDs.

In Table~\ref{tab:qa_curriculum_ablation}, ``Mask-prefix conditioning'' and ``Mask-answer supervision'' yield dataset-dependent grounding changes relative to ``Area-ID supervision'': AP@0.5 increases on CoPESD but decreases when averaged across the two EndoVis datasets. The two mask variants also increase mIoU on CoPESD but decrease the EndoVis-average mIoU. The ``Mask curriculum'' configuration has higher EndoVis-average AP@0.5 than the first four configurations; its CoPESD AP@0.5 is lower than that of ``Mask-answer supervision'' and higher than those of the other three. Our full model achieves the highest short-phrase accuracy, AP@0.5, and mIoU across the reported dataset columns. Table~\ref{tab:qa_area_grounding} further shows that the full configuration achieves the highest area-ID accuracy, Area AP50, and Direct AP50 on all three datasets.

With the full training configuration, Direct AP50 exceeds Area AP50 on all three datasets, including 68.2 versus 48.6 on EndoVis-18 and 80.7 versus 52.3 on CoPESD. These results support using the selected slot areas as spatial references while allowing the model to predict object-specific boxes that differ from the boxes enclosing those areas.

\begin{figure}[t]
 \centering
 \includegraphics[width=\columnwidth]{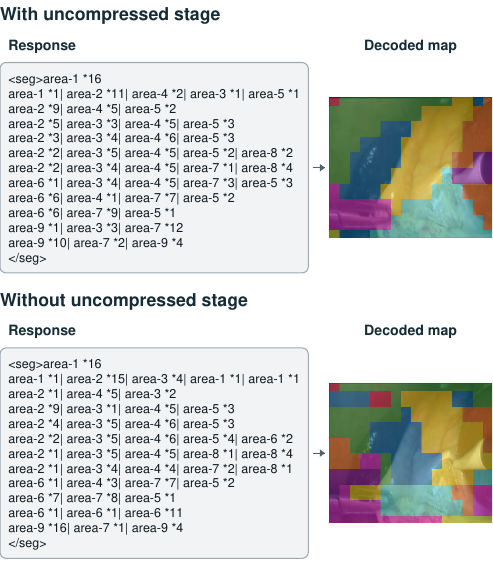}
 \caption{\textbf{Intermediate slot-mask prediction with and without an earlier uncompressed-mask stage.} For the same image and compressed-mask query, generated descriptions are reverse-decoded into area maps. These curriculum-task outputs visualize the spatial organization recovered from the compact slot prefix.}
 \label{fig:qualitative-compression-ablation}
\end{figure}
\noindent\textbf{Intermediate Mask-Prediction Diagnostics.}\label{append:uncompressed-descriptor}
Our intermediate mask-prediction task adapts the semantic-descriptor
representation of Text4Seg~\cite{lan2025text4seg} to slot--language
grounding. Text4Seg constructs mask-description targets from annotated
segmentation masks and predicts them using the visual-token interface
of its underlying MLLM. In our method, the mask targets instead arise from SA decomposition, and the LLM predicts their textual
descriptions from a compact, area-labeled slot prefix. This requires
recovering spatial assignments from compressed region representations
and associating them with image-specific area IDs, motivating the
progression from uncompressed to compressed mask supervision.
Figure~\ref{fig:qualitative-compression-ablation} visualizes this
intermediate learning task through reverse-decoded predictions with
and without the earlier uncompressed-mask stage. With the uncompressed stage, the decoded map forms contiguous regions that follow the instrument tip and the tissue boundary. Without it, the response has the same format but decodes into fragmented, blocky patches.

\begin{table*}[!t]
\centering
\caption{\textbf{Merging-threshold comparison from 0.60 to 0.95.}}
\label{tab:threshold_ablation_low}
\footnotesize
\setlength{\tabcolsep}{0.7pt}
\renewcommand{\arraystretch}{1.12}
\begin{tabularx}{\textwidth}{@{}rr*{14}{>{\raggedleft\arraybackslash}X}@{}}
\toprule
\multirow{2}{*}{$\tau_{\mathrm{inf}}$} & \multirow{2}{*}{\shortstack{\scriptsize \# Visual\\\scriptsize Tokens}} & \multicolumn{4}{c}{\scriptsize Short-Phrase QA (Accuracy)} & \multicolumn{3}{c}{\scriptsize Grounding (AP@0.5)} & \multicolumn{3}{c}{\scriptsize Grounding (mIoU)} & \multicolumn{4}{c}{\scriptsize Visual QA (BLEU)} \\
\cmidrule(lr){3-6}\cmidrule(lr){7-9}\cmidrule(lr){10-12}\cmidrule(lr){13-16}
& & C80 & EV-17 & EV-18 & {\scriptsize CoPESD} & EV-17 & EV-18 & {\scriptsize CoPESD} & EV-17 & EV-18 & {\scriptsize CoPESD} & C80 & EV-17 & EV-18 & {\scriptsize CoPESD} \\
\midrule
0.60 & 1.0 & 66.8 & 33.7 & 48.0 & 70.1 & 2.1 & 10.1 & 10.8 & 9.5 & 17.4 & 21.2 & 92.3 & 18.9 & 43.9 & 41.3 \\
0.65 & 1.0 & 66.9 & 34.5 & 48.4 & 70.2 & 4.7 & 11.7 & 10.4 & 10.1 & 18.1 & 20.9 & 92.3 & 18.9 & 44.4 & 41.3 \\
0.70 & 1.2 & 70.0 & 38.1 & 52.1 & 70.6 & 10.6 & 17.5 & 12.6 & 14.9 & 21.9 & 22.3 & 93.1 & 20.3 & 45.9 & 41.5 \\
0.75 & 1.9 & 80.2 & 47.0 & 60.5 & 72.0 & 25.8 & 32.0 & 22.8 & 25.7 & 32.3 & 29.3 & 95.2 & 24.6 & 49.5 & 43.2 \\
0.80 & 3.6 & 87.7 & 51.1 & 67.5 & 75.3 & 44.9 & 51.5 & 47.6 & 39.2 & 45.6 & 44.6 & 97.0 & 24.7 & \textbf{51.0} & 45.7 \\
0.85 & 6.2 & \underline{90.7} & 54.2 & 70.6 & 77.8 & 56.8 & 62.3 & 70.0 & 46.9 & 53.0 & 58.7 & 97.7 & \underline{25.2} & \underline{50.9} & 46.7 \\
0.90 & 8.9 & \textbf{91.1} & \underline{54.9} & \textbf{71.8} & \underline{78.6} & \underline{59.7} & \underline{66.2} & \underline{79.8} & \underline{49.8} & \underline{55.5} & \underline{64.8} & \underline{97.9} & 24.5 & 50.4 & \textbf{47.2} \\
0.95 & 11.7 & \textbf{91.1} & \textbf{55.7} & \underline{71.5} & \textbf{78.8} & \textbf{61.9} & \textbf{66.5} & \textbf{80.8} & \textbf{50.6} & \textbf{56.3} & \textbf{65.7} & \textbf{98.0} & \textbf{26.4} & 50.5 & \underline{47.0} \\
\bottomrule
\end{tabularx}
\end{table*}

\begin{table}[!t]
\centering
\caption{\textbf{Visual-token counts by dataset (mean $\pm$ standard deviation) for different $\tau$.}}
\label{tab:threshold_visual_tokens_per_dataset}
\footnotesize
\setlength{\tabcolsep}{2pt}
\renewcommand{\arraystretch}{1.12}
\begin{tabular*}{\columnwidth}{@{\extracolsep{\fill}}ccccc@{}}
\toprule
$\tau_{\mathrm{inf}}$ & C80 & EV-17 & EV-18 & CoPESD \\
\midrule
0.60 & $1.00\pm0.01$ & $1.00\pm0.07$ & $1.00\pm0.05$ & $1.00\pm0.00$ \\
0.65 & $1.01\pm0.11$ & $1.03\pm0.18$ & $1.05\pm0.22$ & $1.00\pm0.04$ \\
0.70 & $1.20\pm0.44$ & $1.43\pm0.63$ & $1.43\pm0.62$ & $1.06\pm0.23$ \\
0.75 & $2.12\pm1.02$ & $2.75\pm1.13$ & $2.75\pm1.11$ & $1.47\pm0.65$ \\
0.80 & $4.10\pm1.72$ & $5.56\pm1.69$ & $5.20\pm1.47$ & $2.90\pm1.32$ \\
0.85 & $6.64\pm2.10$ & $8.96\pm1.90$ & $8.26\pm1.67$ & $5.38\pm1.81$ \\
0.90 & $8.75\pm2.03$ & $11.48\pm1.70$ & $10.95\pm1.61$ & $8.48\pm2.09$ \\
0.95 & $10.49\pm2.00$ & $13.06\pm1.58$ & $12.92\pm1.61$ & $12.01\pm1.98$ \\
\bottomrule
\end{tabular*}
\end{table}

\noindent\textbf{Extended Inference-Threshold Comparison.}
Table~\ref{tab:threshold_ablation_low} reports results for inference thresholds \(\tau_{\mathrm{inf}}\) ranging from 0.60 to 0.95 in increments of 0.05, using the checkpoint trained at $\tau_{\mathrm{train}}=0.95$. Grounding evaluates directly predicted boxes using continuous-coordinate IoU, and visual-token counts are question-weighted means of the number of active slot vectors. Bold and underlined values identify the best and second-best distinct scores across all eight rows in this table, including ties after rounding to one decimal place.

Lowering the inference threshold from 0.80 to 0.60 reduces the mean visual-token count from 3.6 to approximately 1.0. Between these endpoints, short-phrase accuracy declines on all four datasets, and grounding mIoU falls from 39.2 to 9.5 on EndoVis-17, from 45.6 to 17.4 on EndoVis-18, and from 44.6 to 21.2 on CoPESD. Among the five settings from 0.60 to 0.80, 0.80 achieves the highest score in every task-metric column, indicating a substantial loss of task information under more aggressive merging for this checkpoint.

Table~\ref{tab:threshold_visual_tokens_per_dataset} reports visual-token counts for C80, EndoVis-17, EndoVis-18, and CoPESD across the same eight inference thresholds. Each question is counted as one observation, so images associated with multiple questions contribute multiple observations. These results show that token-merging behavior varies across datasets and inference thresholds.

\begin{figure*}[t!]
    \centering
    \includegraphics[width=\textwidth,height=0.95\textheight,keepaspectratio]{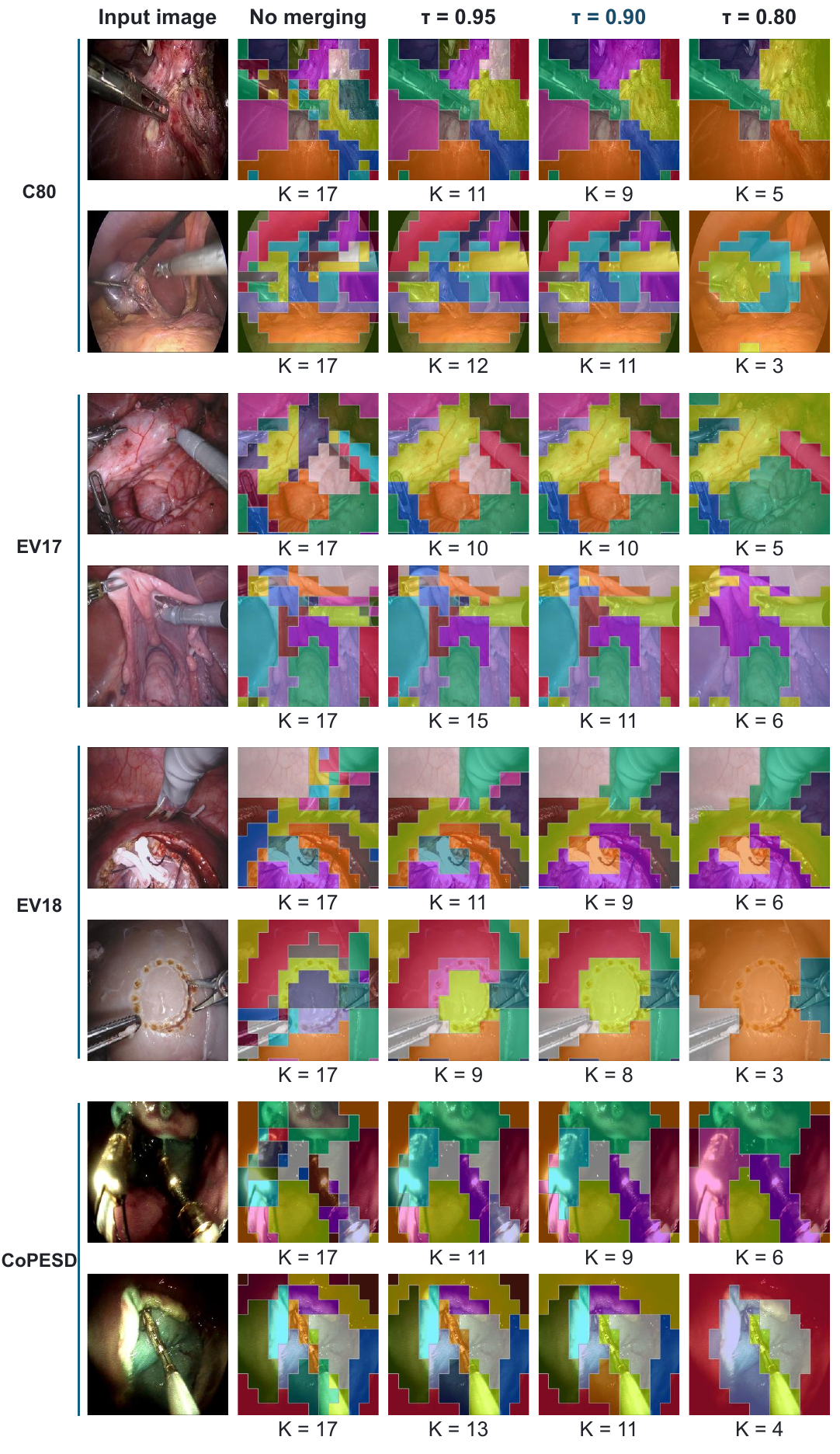}
    \caption{\textbf{Qualitative slot decomposition at different merging thresholds.} In this figure, $K$ denotes the number of active slots, including slots without visible winning pixels.}
    \label{fig:slot-decomposition-thresholds}
\end{figure*}
\noindent\textbf{Qualitative Slot Decomposition.}
Figure~\ref{fig:slot-decomposition-thresholds} compares slot decompositions without merging and at three inference merging thresholds across four datasets. Without merging, all 17 slots remain active, often dividing a single structure into multiple regions. At $\tau=0.95$ and $0.90$, 8--15 slots remain, while instruments and major tissue regions remain distinct. At $\tau=0.80$, only 3--6 slots remain, with tissue and background combined into a few large regions, consistent with the grounding degradation at lower thresholds in Table~\ref{tab:threshold_ablation_low}. These qualitative patterns are consistent with previous observations of Slot Attention pretrained on surgical video~\cite{liao2025future}. The merging threshold thus controls decomposition granularity. Across the examples shown, a single setting of $\tau$  can preserve distinct instrument and tissue regions across all four datasets.

\begin{figure*}[p]
    \centering
    \includegraphics[width=\textwidth,height=0.88\textheight,keepaspectratio]{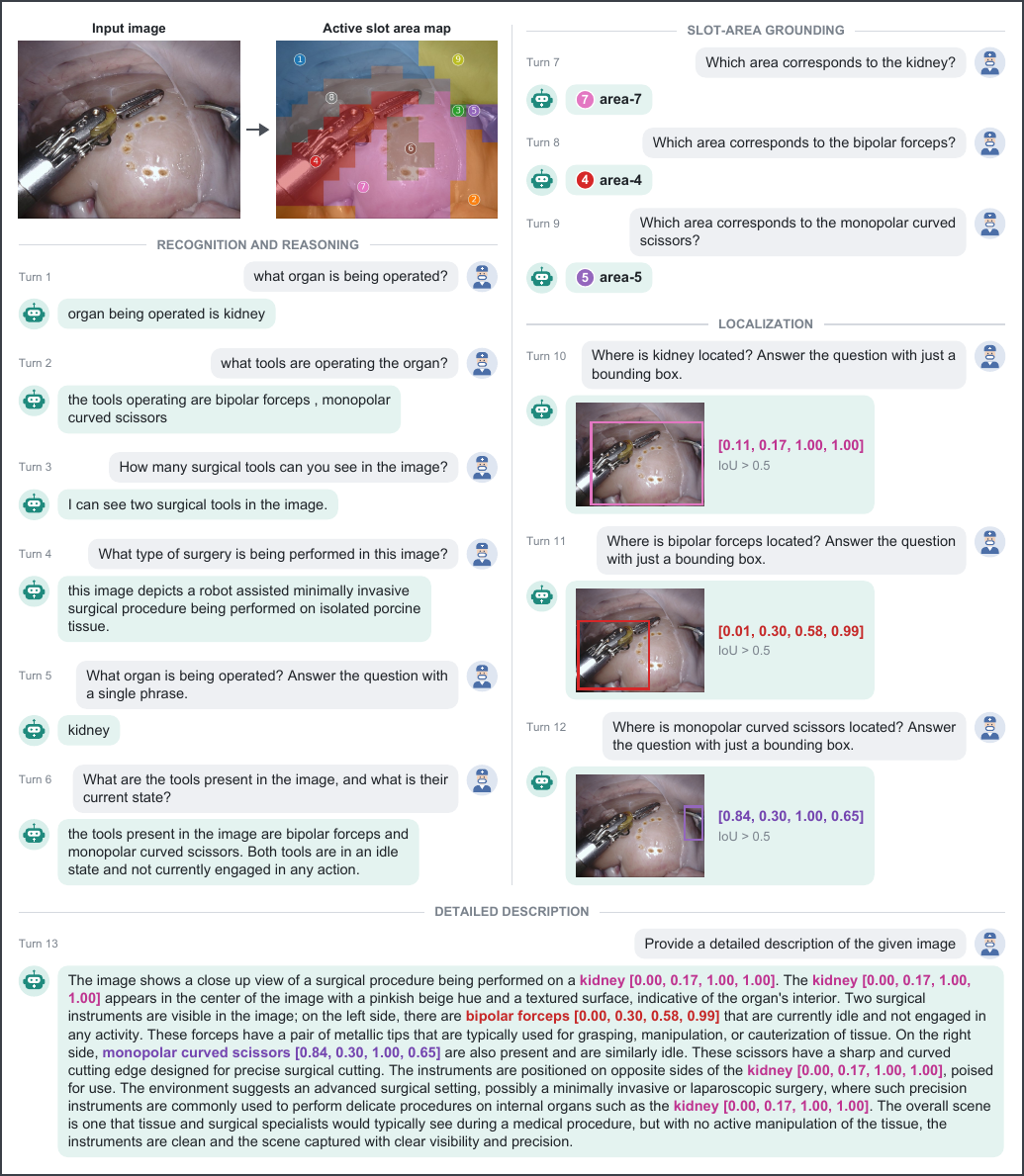}
    \caption{\textbf{Extended multi-turn conversation on a single surgical frame.} The active slot area map shows the image-specific area IDs. Area answers are marked with the color of the corresponding slot region, and each object keeps that color for its predicted box and its mentions in the final description.  }
    \label{fig:extended-multi-turn}
\end{figure*}

\noindent\textbf{Extended Multi-Turn Conversation.}\label{append:extended-multi-turn}
Figure~\ref{fig:extended-multi-turn} illustrates how the model links conversational responses to spatial references across a 13-turn exchange about one surgical frame. It identifies the kidney, bipolar forceps, and monopolar curved scissors, then associates them with area-7, area-4, and area-5, respectively. Subsequent localization responses produce boxes with IoU above 0.5 for all three objects. The final description brings these objects together in a spatially grounded account of the scene. This example demonstrates continuity of object identity across recognition, area grounding, localization, and extended description.

The selected area provides spatial guidance rather than fixing the final bounding box. For example, the predicted kidney box differs slightly from the box enclosing its associated slot area, illustrating that the model can refine localization beyond the area boundary.

\begin{figure*}[p]
    \centering
    \includegraphics[width=\textwidth,height=0.88\textheight,keepaspectratio]{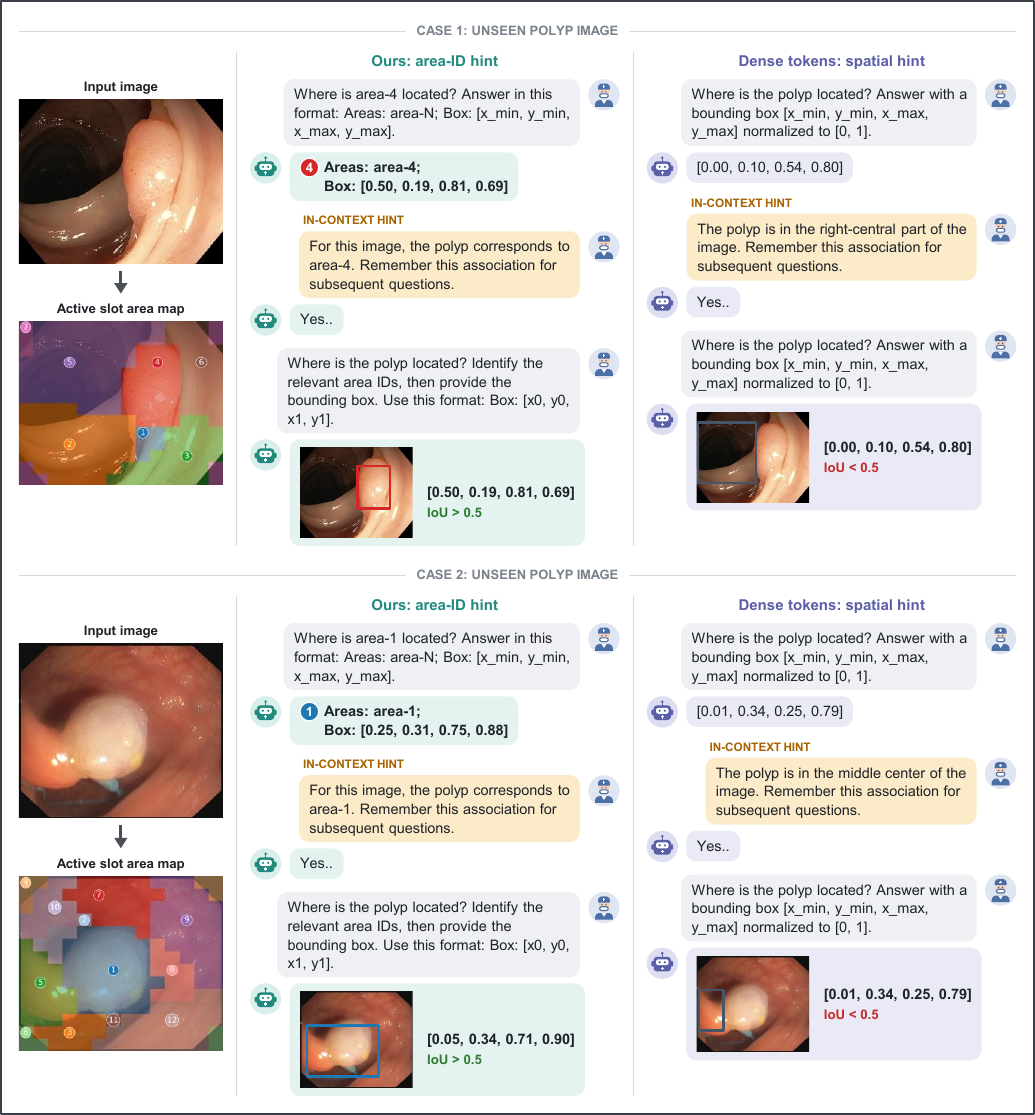}
    \caption{\textbf{In-context hints on unseen polyp images.} Left: input image and active slot area map. Middle: our model receives a hint that links the polyp to an area ID. Right: the dense-token baseline receives a verbal spatial hint. The hint turn is highlighted, and the final answer shows the predicted box on the image.  }
    \label{fig:in-context-polyp}
\end{figure*}

\noindent\textbf{In-Context Area Hints for Interactive Annotation of Unseen Images.}\label{append:in-context-polyp}
Image-specific area IDs provide a simple interface for in-context guidance: a user can associate a target object with a region through a short conversational hint. Figure~\ref{fig:in-context-polyp} illustrates this interaction on two selected colonoscopy polyp images unseen during training. Each conversation has three turns: the user asks for an area's location, associates the polyp with that area ID, and then asks the model to localize the polyp. The dense-token baseline has no area vocabulary, so it is first asked to localize the polyp, then given a verbal spatial hint, and then asked again. After receiving the area hints, our model localizes both polyps, whereas the baseline repeats its initial boxes without predicting correct locations. These examples suggest potential for interactive surgical image annotation, where users guide localization through region references without updating model weights. Extending this interaction to surgical video annotation remains future work.

In the second case, our final box differs from the box reported for area-1, since the area guides the prediction without fixing it. This behavior is consistent with the training formulation, which pairs area references with object-specific bounding-box targets. The results in Table~\ref{tab:qa_area_grounding} complement this observation by showing differences in grounding accuracy between area-derived and directly predicted boxes on the surgical benchmarks.

\begin{figure*}[p]
    \centering
    \includegraphics[width=\textwidth,height=0.90\textheight,keepaspectratio]{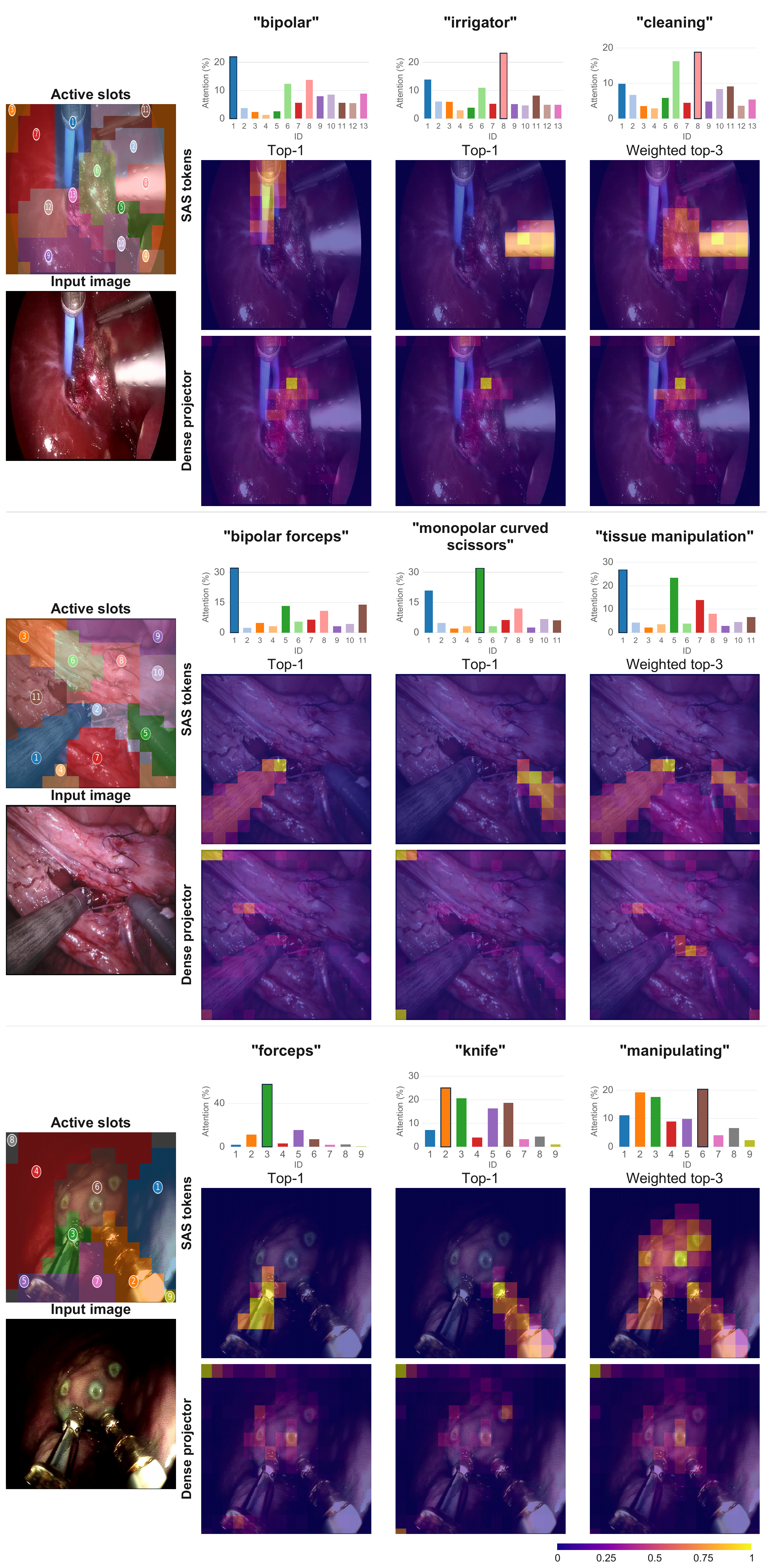}
    \caption{\textbf{Attention from generated language to SAS tokens.} Samples from C80, EndoVis, and CoPESD appear from top to bottom, separated by thin gray lines. Bars show the share of attention assigned to each slot token (\%), by token ID. Each sample shows the active-slot mask and input at left, followed by three quoted phrases with slot-attention bars, the selected slot-based spatial map, and the dense-projector map.}
    \label{fig:attention-supp-1}
\end{figure*}

\begin{figure*}[p]
    \centering
    \includegraphics[width=\textwidth,height=0.90\textheight,keepaspectratio]{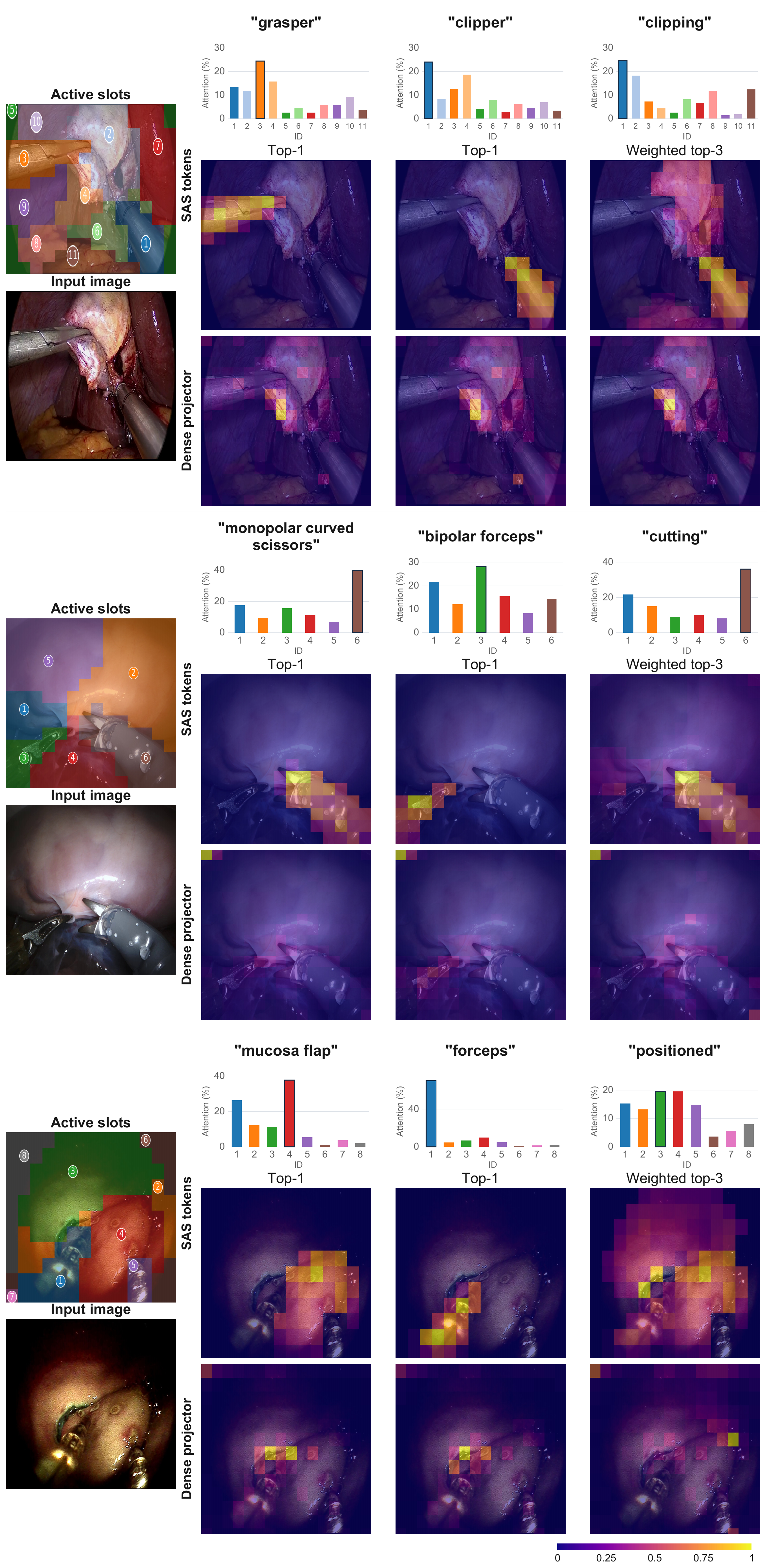}
    \caption{\textbf{Attention from generated language to SAS tokens.} Additional C80, EndoVis, and CoPESD cases, from top to bottom, use the same layout and notation as Figure~\ref{fig:attention-supp-1}.}
    \label{fig:attention-supp-2}
\end{figure*}

\noindent\textbf{Additional Attention Visualizations.}\label{append:phrase-attention}
Figures~\ref{fig:attention-supp-1} and~\ref{fig:attention-supp-2} present two sets of attention visualizations linking generated language to SAS tokens, using examples from C80, EndoVis, and CoPESD.
Each example pairs the active-slot decomposition and input image with attention distributions, slot-based spatial maps, and dense-projector maps. Bar charts shows the percentage of attention assigned to each slot token, normalized across slots to sum to 100\%. The x-axis label ``ID'' denotes the token ID. The outlined bar identifies the most-attended slot. Token IDs and colors are local to each image decomposition. Top-1 panels show patch-level attention for the most-attended slot. Weighted top-3 panels combine the patch-level attention maps of the three most-attended slots.

Instrument terms are visualized using top-1 attention maps. Action terms such as ``cleaning'' and ``manipulating'' involve attention across multiple slots and are therefore visualized using weighted top-3 maps.

Attention maps for instrument terms such as ``bipolar forceps'', ``knife'', and ``grasper'' highlight the corresponding instrument regions, revealing word-to-region correspondence through the selected slots. Maps for action terms such as ``cleaning'', ``clipping'', and ``manipulating'' highlight multiple participating regions, including the surgical instruments and target tissue. The combined maps thus capture the spatial context of the interaction beyond a single named object. In these examples, the dense-projector maps often concentrate on isolated patches or show similar responses across different terms, with less coherent coverage of the relevant objects. The slot-based maps provide a clearer qualitative association between object terms and their regions, and between action terms and the regions involved in the interaction.
 
\end{document}